\pdfoutput=1

\documentclass[11pt]{article}

\usepackage[]{ACL2023}
\usepackage{times}
\usepackage{latexsym}

\usepackage[T1]{fontenc}

\usepackage[utf8]{inputenc}

\usepackage{microtype}

\usepackage{inconsolata}
\usepackage{soul}
\usepackage{url}
\usepackage{caption}
\usepackage{graphicx}
\usepackage{amsthm,amsmath,amssymb,mathtools}
\usepackage{diagbox}

\usepackage{hyperref}
\usepackage{url}

\usepackage{academicons}
\usepackage{fontawesome5}

\usepackage{soul}
\usepackage{url}
\usepackage{caption}
\usepackage{graphicx}
\usepackage{amsthm,amsmath,amssymb,mathtools}
\usepackage{diagbox}

\usepackage{newfloat}
\usepackage{listings}
\usepackage{subfigure}
\usepackage{hhline,booktabs,colortbl,multirow,tabularx,diagbox,tablefootnote,threeparttable} 
\usepackage{multirow}

\usepackage{prettyref}
\usepackage{xcolor}
\usepackage[shortlabels]{enumitem}
\usepackage{float}
\usepackage{wrapfig}
\usepackage{natbib}

\definecolor{mycyan}{gray}{.7}

\newrefformat{fig}{Figure~\ref{#1}}
\newrefformat{tab}{Table~\ref{#1}}
\newrefformat{sec}{Section~\ref{#1}}
\newrefformat{app}{Appendix~\ref{#1}}
\newrefformat{alg}{Algorithm~\ref{#1}}
\newrefformat{property}{Property~\ref{#1}}
\newrefformat{theorem}{Theorem~\ref{#1}}
\newrefformat{corollary}{Corollary~\ref{#1}}
\newrefformat{proposition}{Proposition~\ref{#1}}
\newrefformat{def}{Definition~\ref{#1}}
\newrefformat{eq}{equation~(\ref{#1})}
\newrefformat{app}{Appendix~\ref{#1}}
\newcommand{\pref}{\prettyref}

\def\our{\textsc{Rapid}}
\def\pd{\textsc{PD}}

\def\disp{\textsc{DISP}}
\def\fgws{\textsc{FGWS}}
\def\rsv{\textsc{RS\&V}}

\def\bert{\textsc{BERT}}
\def\deberta{\textsc{DeBERTa}}

\def\bae{\textsc{BAE}}
\def\pwws{\textsc{PWWS}}
\def\textfooler{\textsc{TextFooler}}

\def\pso{\textsc{PSO}}
\def\iga{\textsc{IGA}}

\def\wordbugger{\textsc{DeepWordBug}}
\def\clare{\textsc{Clare}}

\def\sst2{\texttt{SST2}}
\def\amazon{\texttt{Amazon}}
\def\agnews{\texttt{AGNews}}
\def\yahoo{\texttt{Yahoo!}}

\def\pa{\texttt{Phase \#1}}
\def\pb{\texttt{Phase \#2}}

\newtheorem{remark}{Remark}

\title{The Best Defense is Attack: Repairing Semantics in Textual Adversarial Examples}

\author{
	Heng Yang$^1$, Ke Li$^1$ \\ 
	$^1$Department of Computer Science, University of Exeter, EX4 4QF, Exeter, UK \\
	\texttt{\{hy345, k.li\}@exeter.ac.uk} \\
}
\begin{document}
\maketitle
\begin{abstract}

Recent studies have revealed the vulnerability of pre-trained language models to adversarial attacks. Existing adversarial defense techniques attempt to reconstruct adversarial examples within feature or text spaces. However, these methods struggle to effectively repair the semantics in adversarial examples, resulting in unsatisfactory performance and limiting their practical utility. To repair the semantics in adversarial examples, we introduce a novel approach named Reactive Perturbation Defocusing (\our). \our\ employs an adversarial detector to identify fake labels of adversarial examples and leverage adversarial attackers to repair the semantics in adversarial examples. Our extensive experimental results conducted on four public datasets, convincingly demonstrate the effectiveness of \our in various adversarial attack scenarios. To address the problem of defense performance validation in previous works, we provide a demonstration of adversarial detection and repair based on our work, which can be easily evaluated at \url{https://tinyurl.com/22ercuf8}.

\end{abstract}


\section{Introduction}
\label{sec:introduction}

\begin{figure}[t!]
    \centering
    \includegraphics[width=\linewidth]{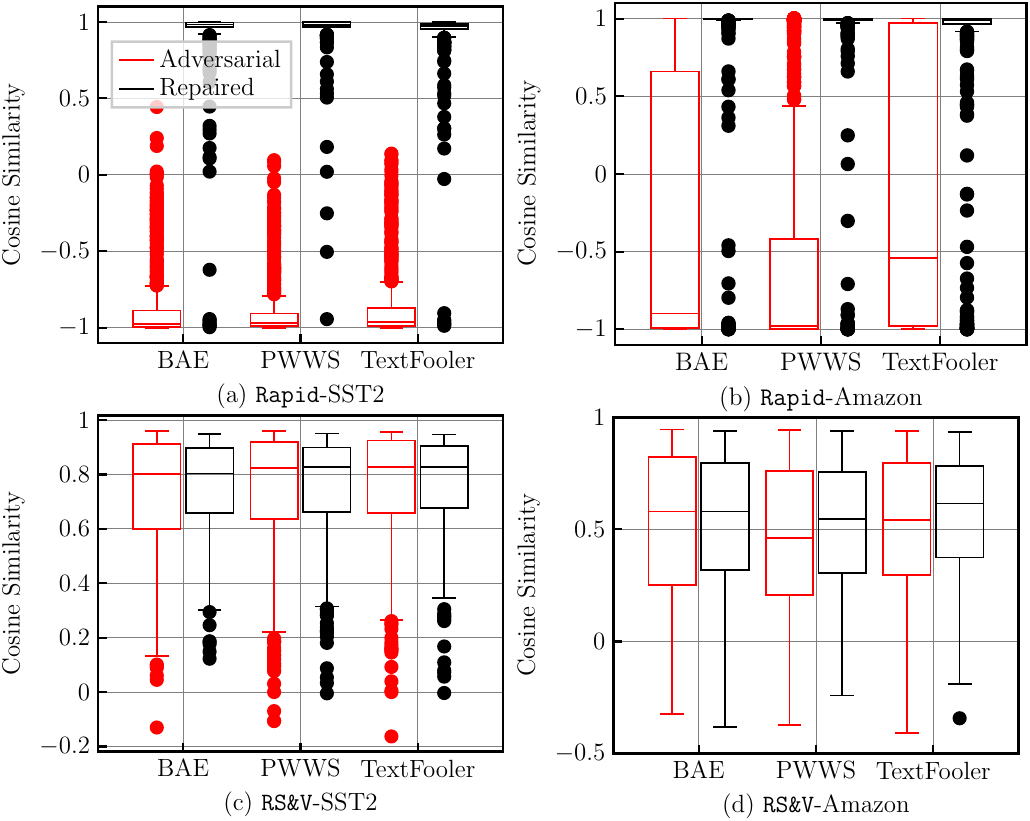}
    \caption{Box plots of the cosine similarity between the \textit{adversary--natural example pairs} (marked in \textcolor{red}{red}) and the \textit{repaired adversary--natural example pairs} obtained by \our\ versus \rsv. The cosine similarity is evaluated based on the features extracted by the victim models of \our\ and \rsv, respectively. The larger the cosine similarity, the more similar the corresponding example pair. It is observed that the victim model cannot discern the semantic differences between the adversaries and the repaired adversaries produced by \rsv, whereas \our\ can precisely differentiate between adversaries and natural examples. Conversely, when using \our, the repaired adversaries regain their semantic alignment with the natural examples.}
    \label{fig:intro}
    \vspace{-15pt}
\end{figure}

Pre-trained language models (PLMs) have achieved state-of-the-art (SOTA) performance across a variety of natural language processing tasks~\citep{WangPNSMHLB19,WangSMHLB19}. However, PLMs are reported to be highly vulnerable to adversarial examples, a.k.a., \textit{adversaries}~\citep{LiJDLW19,GargR20,LiMGXQ20,JinJZS20,LiZPCBSD21,BoucherSAP22}, created by subtly altering selected words in natural examples, a.k.a. \textit{clean} or \textit{benign examples}~\citep{MorrisLYGJQ20}. While the significance of textual adversarial robustness regarding adversarial attacks has broadly recognized within the deep learning community~\cite{AlzantotSEHSC18,RenDHC19,ZangQYLZLS20,ZhangZC020,JinJZS20,LiZPCBSD21,WangXLCL22,XuSGGH23}, efforts to enhance adversarial robustness remain very limited, especially when comparing to other deep learning fields like computer vision~\citep{RonyHOASG19,GowalRWSCM21,WangPDLLY23,XuSGGH23}. Current works on textual adversarial robustness can be classified into three categories---\textit{adversarial defense}, \textit{adversarial training}~\citep{LiuZWLC20,LiuLYLSLS20,IvgiB21,DongLLYZ21,DongLJ021}, and \textit{adversary reconstruction}~\citep{ZhouJCW19,JonesJRL20,BaoWZ21,KellerME21,MozesSKG21,LiSZMQ22,ShenZJC23}. Since both adversarial training and reconstruction are resource-intensive, there has been growing interest in adversarial defense. Nevertheless, the current adversarial defense techniques have two bottlenecks.
\begin{itemize}[nosep,noitemsep,nolistsep]
    \item[\faCarCrash] Current works can hardly identify the semantic discrepancies between natural and adversarial examples\footnote{In this work, we refer to the semantics in adversaries as the features encoded by PLM for simplicity.}. Let us use \rsv, a recent adversarial defense~\citep{WangXH22}, as an example. As shown in~\pref{fig:intro}, it is clear that \rsv\ fails to discern the semantic differences between adversarial and repaired examples. This is attributed to the augmentation method used in \rsv\ that is not only untargeted but also does not effectively identify and neutralize adversaries.

    \item[\faCarCrash] Given the time-intensive nature of the defense process, adversarial defense is also notorious for its computational inefficiency~\citep{MozesSKG21,WangXH22}. This can be partially attributed to their inability to \textit{pre-detect} adversaries and indiscriminately process all input texts. This not only wastes computational budget on unnecessary defense actions regarding natural examples, but also leads to an unwarranted defensive stance towards natural examples, which may further compromise performance.
\end{itemize}

\begin{figure*}[t!]
    \centering
    \includegraphics[width=\linewidth]{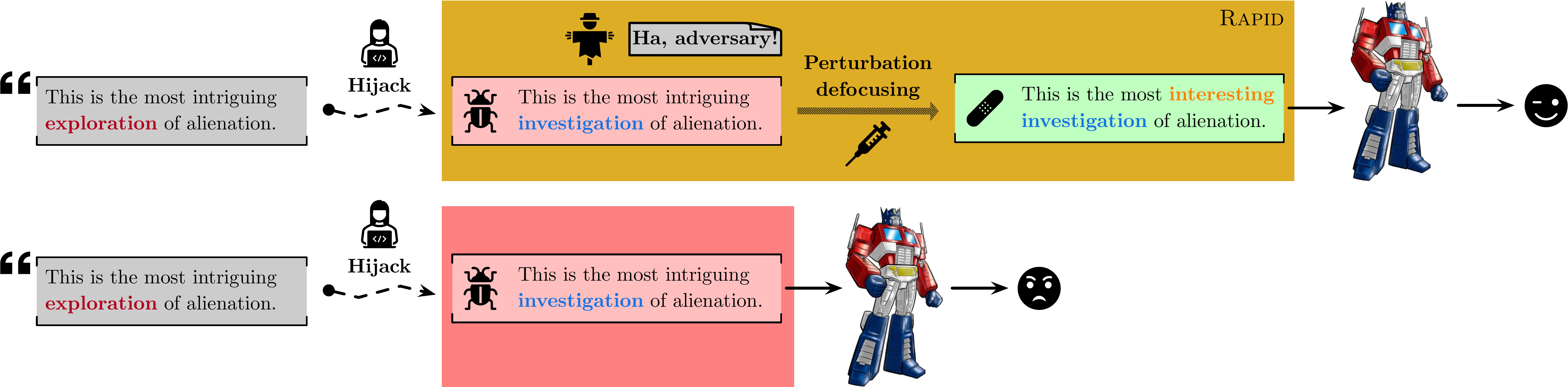}
    \caption{A pedagogical example of \our\ in sentiment analysis. The original word in this example is \textcolor{red}{exploration}. Perturbation defocusing repairs the adversary by injecting perturbations (\textcolor{orange}{interesting}) to distract the objective model from the malicious perturbation (i.e., \textcolor{blue}{investigation}). \our\ only implements defense on the pre-detected adversary.} 
    \label{fig:pd}
    \vspace{-15pt}
\end{figure*}

Bearing the above two challenges in mind, we propose a simple yet effective textual adversary defense paradigm, named reactive perturbation defocusing (\our), which has the following two distinctive features.
\begin{itemize}[nosep,noitemsep,nolistsep]
    \item[\faLightbulb] To address the first bottleneck, we propose a novel concept of perturbation defocusing (\pref{sec:pd}). The basic idea is to leverage adversarial attackers to \textit{re-inject} some perturbations into the \textit{pre-detected} adversaries to distract the victim model from malicious perturbations, and to \textit{repair} these adversaries based on the inherent robustness of the victim models. Further, the accuracy of adversarial defense is augmented by a pseudo-semantic similarity filtering strategy (\pref{sec:similarity}).

    \item[\faLightbulb] To overcome the second bottleneck, \our\ trains an \textit{in-victim-model} adversarial detector, without introducing additional cost (\pref{sec:phase_1}), to proactively concentrate the defense efforts on the examples \textit{pre-detected} as adversaries. In particular, this adversarial detector is jointly trained with the victim model in a multi-task way, and is capable of recognizing adversaries generated by different attackers. This helps not only minimize collateral impacts on natural examples~\citep{XuZZLHCH22}, but also reduces the waste of computational budget upon defending against natural examples. 

\end{itemize}
\pref{fig:pd} provides a pedagogical example of the working mechanism of \our\ in the context of sentiment analysis. There are four key takeaways from our empirical study.
\begin{itemize}[nosep,noitemsep,nolistsep]
    \item[\faMagic] \our\ achieves up to $99.9\%$ repair accuracy upon pre-detected adversaries, significantly surpassing text/feature-level reconstruction and voting-based methods (\pref{tab:main}).

    \item[\faMagic] \our\ reduces nearly $50\%$ computational cost for adversarial defense compared against adversarial attack (\pref{tab:efficiency2}).

    \item[\faMagic] \our\ is robust in recognizing and defending against a wide range of unknown adversarial attacks (\pref{tab:rq1}), such as \clare~\citep{LiZPCBSD21} and large language models like \texttt{ChatGPT-3.5}~\citep{OpenAI23}.

    \item[\faMagic] We develop a user-friendly API\footnote{For the sake of anonymous requirement, we promise to release this tool upon the acceptance of this paper.} as a benchmarking platform for different adversarial attackers under the defense of \our.
\end{itemize}

\section{Proposed Method}
\label{sec:proposed_method}

Our proposed \our\ framework comprises two phases. \underline{\pa} trains a joint model that not only performs the standard text classification task but is also capable of detecting adversaries. \underline{\pb} is dedicated to implementing pseudo-supervised adversary defense based on \pd. It diverts the victim model's attention from malicious perturbations, and rectifies the outputs without compromising performance on natural examples. 

\subsection{\pa: joint model training}
\label{sec:phase_1}

The crux of \pa\ is the joint training of two models: one is the victim model as the standard text classifier, and the other is an in-victim-model adversarial detector, which is a binary classifier that pre-detect adversaries before the defense.

\subsubsection{Multi-attack-based adversary sampling}
\label{sec:sampling}

To derive the data used for training the adversarial detector, we apply adversarial attack methods upon the victim model $F_S$ to sample adversaries. To enable the adversarial detector to identify various unknown adversaries, we employ three widely used open-source adversarial attackers: \bae~\citep{GargR20}, \pwws~\citep{RenDHC19}, and \textfooler~\citep{JinJZS20}. For each data instance $\langle\mathbf{x},y\rangle\in\mathcal{D}$, the set of natural examples, we apply each of the adversarial attackers to sample three adversaries\footnote{The formulation of word-level adversarial attack is available in \pref{app:preliminary}.}:
\begin{equation}
    \langle\tilde{\mathbf{x}},\tilde{y}\rangle_i\leftarrow\mathcal{A}_i\left(F_S,\langle\mathbf{x},y\rangle\right),
\end{equation}
where $\mathcal{A}_i$, $i\in \{1,2,3\}$, represents \bae, \pwws, and \textfooler, respectively. $\langle\tilde{\mathbf{x}},\tilde{y}\rangle_i$ is the adversary generated by $\mathcal{A}_i$. Note that we collect all adversaries, including both successful and failed ones, to constitute the adversarial dataset $\tilde{\mathcal{D}}$. Finally, we compose a hybrid dataset as shown in the left part of \pref{fig:workflow}. $\overline{\mathcal{D}}:=\mathcal{D}\bigcup\tilde{\mathcal{D}}$ for the joint model training.

\subsubsection{Joint model training objectives}
\label{sec:joint_training}

To conduct the joint model training of both the victim model and the adversarial detector, we propose an aggregated loss function as follows:
\begin{equation}
    \mathcal{L}:=\mathcal{L}_\mathrm{c}+\mathcal{L}_\mathrm{d}+\mathcal{L}_\mathrm{a}+\lambda||\boldsymbol{\theta}||^2_2,
	\label{eq:loss_function}
\end{equation}
where $\lambda$ is the $\ell_2$ regularization parameter, and $\boldsymbol{\theta}$ represents the parameters of the underlying PLM. $\mathcal{L}_\mathrm{c}$, $\mathcal{L}_\mathrm{d}$, and $\mathcal{L}_\mathrm{a}$ denotes the loss for training a standard classifier, an adversarial detector, and adversarial training, respectively.
\begin{itemize}[nosep,noitemsep,nolistsep]
    \item\underline{\textit{Standard classification loss} $\mathcal{L}_\mathrm{c}$}: Here we use the cross-entropy loss widely used for text classification:
        \begin{equation}
            \mathcal{L}_\mathrm{c}:=-\sum_{i=1}^C\left[p_i\log\left(\tilde{p}_i\right)+q_i\log\left(\tilde{q}_i\right)\right],
        \end{equation}
        where $C$ is the number of classes. $p$ and $\tilde{p}$ respectively indicate the true and predicted probability distributions of the standard classification label, while $q$ and $\hat{q}$ represent any incorrect standard classification label and its likelihood, respectively. Note that the labels of the adversaries within $\overline{\mathcal{D}}$ are set to a dummy value $\varnothing$ in this loss. By doing so, we can make sure that $\mathcal{L}_\mathrm{c}$ focuses on the natural examples.

    \item\underline{\textit{Adversarial detection loss $\mathcal{L}_\mathrm{d}$}}: It only calculates the binary cross-entropy for both natural examples and adversaries within $\overline{\mathcal{D}}$, where the labels are either $0$ or $1$ in practice. Note that $\mathcal{L}_\mathrm{d}$ is used to train the adversarial detector as a binary classifier that determines whether the input example is an adversary or not.

    \item\underline{\textit{Adversarial training loss $\mathcal{L}_\mathrm{a}$}}: In practice, the calculation of $\mathcal{L}_\mathrm{a}$ is the same as $\mathcal{L}_\mathrm{c}$. To improve the robustness of adversaries, $\mathcal{L}_\mathrm{a}$ only calculates the loss for the adversaries by setting the labels of natural examples within $\overline{\mathcal{D}}$ as a dummy $\varnothing$. By doing so, we can prevent this adversarial training loss from negatively impacting the performance on pure natural examples, which have been reported to be notorious in recent studies~\citep{DongLJ021,DongLLYZ21}.
\end{itemize}
All in all, each instance $\langle\overline{\mathbf{x}},\overline{\mathbf{y}}\rangle\in\overline{\mathcal{D}}$ is augmented with three different labels to accommodate these three training losses, where $\overline{\mathbf{y}}:=(\overline{y}_1,\overline{y}_2,\overline{y}_3)^\top$.

\subsection{\pb: reactive adversarial defense}
\label{sec:defense}

\begin{figure*}[t!] 
	\centering
	\includegraphics[width=.6\linewidth]{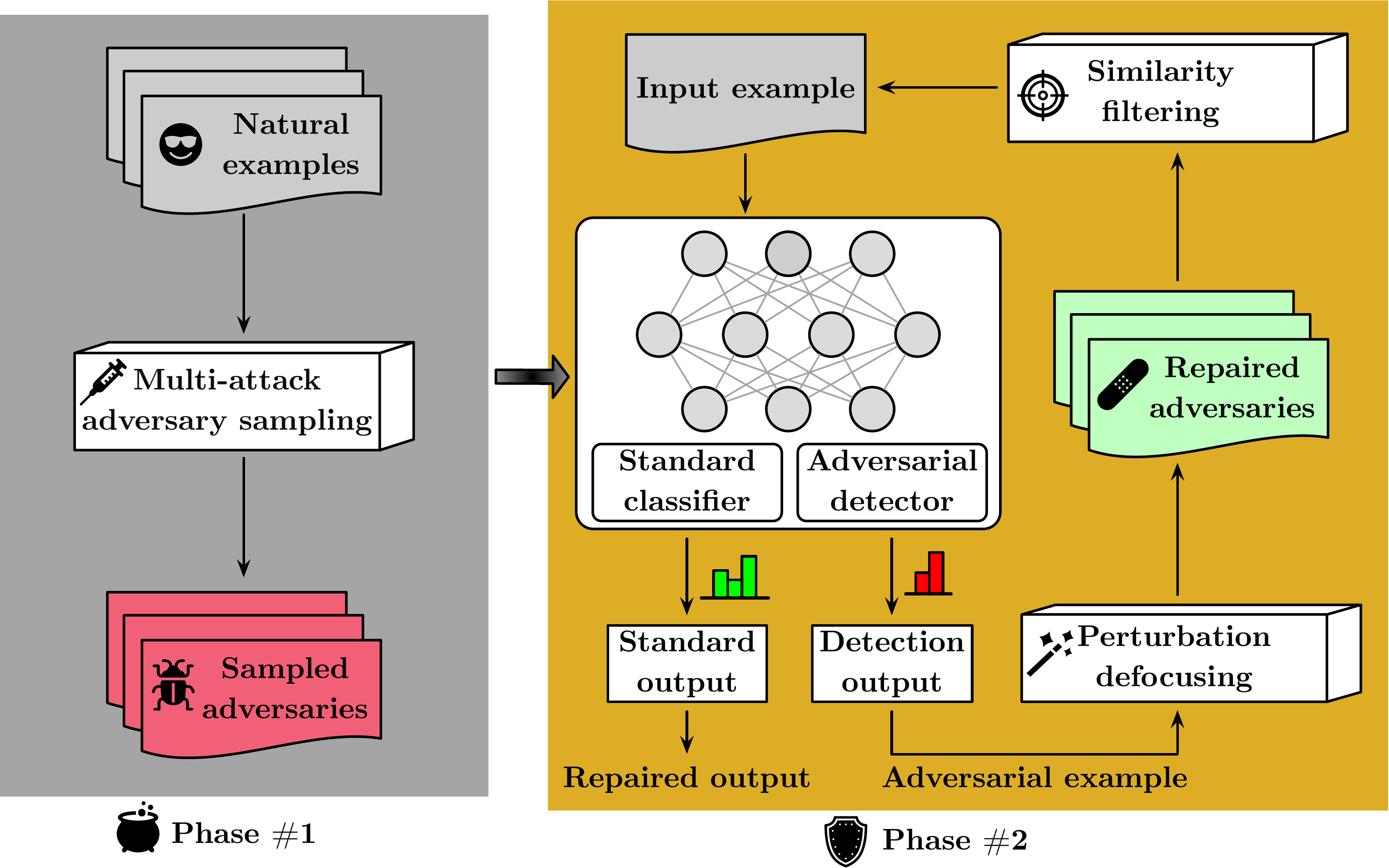}
	\caption{The overall architecture and workflow of \our.}
	\label{fig:workflow}
    \vspace{-15pt}
\end{figure*}

To address the efficiency and semantic challenges discussed in~\pref{sec:introduction}, the reactive adversarial defense consists of the following three steps.

\subsubsection{Adversarial defense detection}
\label{sec:add}

Our preliminary experiments suggested that PLMs like \bert\ and \deberta\ are sensitive to semantic shifts caused by adversarial attacks. Thereby, different from the current adversarial defense methods, which often indiscriminately run defense upon all input examples, we will first apply the joint model $F_J$ trained in the \pa\ to determine whether the input $\hat{\mathbf{x}}$ is adversarial or not using the following prediction:
\begin{equation}
	(\hat{y}_1, \hat{y}_2, \hat{y}_3) \leftarrow F_J(\hat{\mathbf{x}}) ,
	\label{eq:prediction}
\end{equation}
where $\hat{y}_1$, $\hat{y}_2$, and $\hat{y}_3$ are predicted labels according to the three training losses in~\pref{eq:loss_function}, respectively. Thereafter, only the inputs identified as adversaries (i.e., those with $\hat{y}_2=1$) are used for the follow-up perturbation defocusing.

\subsubsection{Perturbation defocusing}
\label{sec:pd}

The basic idea of this perturbation defocusing is to inject \textit{safe} perturbations into the adversary $\hat{\mathbf{x}}$ identified by the adversarial defense detection in~\pref{sec:add}. The process is shown in \texttt{Phase \#2} in\pref{fig:workflow}. In practice, we apply an adversarial attacker to \textit{attack} $\hat{\mathbf{x}}$ to obtain a \textit{repaired example}:
\begin{equation}
    \langle\tilde{\mathbf{x}}^\mathrm{r},\tilde{y}_1^\mathrm{r}\rangle\leftarrow\hat{\mathcal{A}}_{PD}\left(F_J,\langle\hat{\mathbf{x}},\hat{y}_1\rangle\right),
    \label{eq:rpd}
\end{equation}
where $\hat{y}_1$ is the predicted label of $\hat{\mathbf{x}}$, and $\hat{\mathcal{A}}_{PD}$ is an adversarial attacker\footnote{We choose \pwws\ because it is cost-effective, and it can be replaced by any (or an ensemble of) adversarial attackers.}. Note that the above perturbation is considered safe because it does not alter the semantics of $\hat{\mathbf{x}}$. By this means, we divert the standard classifier's focus away from the malicious perturbations, allowing the standard classifier to concentrate on the adversary's original semantics. In essence, the repaired examples can be correctly classified based on their own robustness.

\subsubsection{Pseudo-semantic similarity filtering}
\label{sec:similarity}

Last but not least, to prevent repaired adversaries from being misclassified, we propose a feature-level pseudo-semantic similarity filtering strategy to mitigate semantic bias. Specifically, for each $\hat{\mathbf{x}}$, we generate a set of repaired examples $\mathcal{S}:=\{\tilde{\mathbf{x}}^\mathrm{r}_i\}_{i=1}^k$. Then, we encode these repaired examples using $F_J$ to extract their semantic features. Thereafter, for each repaired example within $\mathcal{S}$, we calculate its similarity score as:
\begin{equation}
    s_i=\frac{\sum_{j=1,j\neq i}^k \mathrm{sim}{\left(\mathcal{H}_i,\mathcal{H}_j\right)}}{k},
\end{equation}
where $\mathcal{H}_i$ and $\mathcal{H}_j$ are the hidden states of $\tilde{\mathbf{x}}^\mathrm{r}_i$ and $\tilde{\mathbf{x}}^\mathrm{r}_j$ encoded by $F_J$, and $\mathrm{sim}(\ast,\ast)$ evaluates the cosine similarity. For the sake of efficiency, we set $k=3$ in this paper. After the defense, the label of the repaired $\hat{\mathbf{x}}$ is assigned as the predicted label of the repaired example within $\mathcal{S}$ having the largest similarity score.

\begin{remark}
    Generally speaking, the basic idea of an adversarial attacker is to inject some (usually limited) malicious perturbations into a natural example, thus fooling the victim model. This often results in adversaries looking similar to the natural examples. However, the corresponding semantics are often \lq destroyed\rq\ after the perturbation. This inspires us to introduce a new adversarial attacker, even though different from the malicious attacker, to attack and thus repair the malicious semantics of adversaries provided that we know its fake labels. Further, due to the principle of minimizing edits when changing the prediction, we can also mitigate text space shifts in repaired examples along with the semantics.
\end{remark}

\begin{remark}
    Note that the defender in \our\ is decoupled with the adversarial detector, and its performance is agnostic to the adversarial attackers used for this adversary sampling. The empirical results in~\pref{tab:rq1} demonstrate that the adversarial detector can adapt to unknown attack methods, even when trained on a small set of adversaries.
\end{remark}


\vspace{-.5em}
\section{Experimental Settings}
\label{sec:settings}
\vspace{-.5em}

In this section, we introduce the experimental settings used in our experiments.

\begin{table}[ht]
	\centering
	\caption{The statistics of datasets used for evaluating \our. We use subsets from \amazon, \agnews~and \yahoo~ datasets to evaluate \our~ as the previous works due to high resource occupation. }
	\resizebox{\linewidth}{!}{
		\renewcommand{\arraystretch}{1.2}
		\begin{tabular}{c|c|c|c|c}
			\hline
			\multirow{2}[1]{*}{\textsc{Dataset}} & \multirow{2}[1]{*}{\textsc{Categories}} & \multicolumn{3}{c}{\textsc{Number of Examples}} \\
            \cline{3-5}          &       & \textsc{Training} & \textsc{Valid} & \textsc{Testing} \\
			\hline
			\sst2  & $2$     & $6,920$  & $872$   & $1,821$ \\
			\hline
			\amazon & $2$     & $7,000$  & $1,000$  & $2,000$ \\
			\hline
			\agnews & $4$     & $120,000$  & $0$  & $7,600$ \\
			\hline
			\yahoo & $10$     & $1,400,000$  & $0$  & $60,000$ \\
			\hline
		\end{tabular}%
	}
	\label{tab:datasets}%
        \vspace{-10pt}
\end{table}%

\noindent
\textbf{Victim models:} while any PLM can be used in a plug-in manner in \our, this paper considers \bert~\citep{DevlinCLT19} and \deberta~\citep{HeGC21}, two widely used PLMs based on the transformer structure\footnote{\url{https://github.com/huggingface/transformers}}, as both the victim classifier and the joint model. Their corresponding hyperparameter settings are in~\pref{app:hyperparameter}.

\noindent
\textbf{Datasets:} we consider three widely used text classification datasets\footnote{We have released the detailed source codes and processed datasets in the supplementary materials.}, including \sst2~\citep{SocherPWCMNP13}, \amazon~\citep{ZhangZL15}, and \agnews~\citep{ZhangZL15} whose key statistics are outlined in~\pref{tab:datasets}. \sst2 and \amazon\ are binary sentiment classification datasets. \agnews\ and \yahoo\ is a multi-categorical news classification dataset containing $4$ and $10$ categories, respectively.

\noindent
\textbf{Adversarial attackers:} our experiments employ three open-source attackers provided by \textsc{TextAttack}\footnote{\url{https://github.com/QData/TextAttack}}~\citep{MorrisLYGJQ20}. Their functionalities are outlined as follows, while their working mechanisms are in~\pref{app:sampling-attackers}.

\begin{enumerate}[leftmargin=*,label=\alph*),nosep,noitemsep,nolistsep]
    \item \underline{\textit{Adversary sampling}}. \bae, \pwws\ and \textfooler\ are used to sample adversaries for training the adversarial detector (\pref{sec:phase_1}). Since they represent different types of attacks, we can train a detector that recognizes a variety of adversarial attacks.

    \item \underline{\textit{Adversary repair}}. We employ \pwws\ as the attacker $\hat{\mathcal{A}}_{PD}$ in the perturbation defocusing (\pref{sec:defense}). Compared to \bae, our preliminary experiments demonstrate that \pwws\ rarely changes the natural examples' semantics, and it is more computationally efficient than \textfooler.

    \item \underline{\textit{Generalizability evaluation}}. We use \iga~\citep{WangH0021}, \wordbugger~\citep{GaoLSQ18}, \pso~\citep{ZangQYLZLS20} and \clare\ to evaulate \our's generalization capability.
\end{enumerate}

\noindent
\textbf{Evaluation metrics:} we use the following five fine-grained metrics\footnote{The mathematical definitions of these evaluation metrics can be found in~\pref{app:metrics}.} for text classification to evaluate the adversarial defense performance.
\begin{itemize}[nosep,noitemsep,nolistsep]
    \item\underline{\textit{Nature accuracy} (\textsc{NtA})}: it evaluates the victim's performance on the target dataset that only contains natural examples.

    \item\underline{\textit{Attack accuracy} (\textsc{AtA})}: It evaluates the victim's performance under adversarial attacks.

    \item\underline{\textit{Detection accuracy} (\textsc{DtA})}: It measures the defender's adversaries detection performance. 

    \item\underline{\textit{Defense accuracy} (\textsc{DfA})}: It evaluates the defender's performance of adversaries repair. 

    \item \underline{\textit{Repaired accuracy} (\textsc{RpA})}: It evaluates the victim's performance on the attacked dataset after being repaired.
\end{itemize}
Note that we evaluate the adversarial detection and defense performance on the entire testing set, while current works~\citep{XuZZLHCH22,YangWH22,DongLJ021,DongLLYZ21} only evaluated a small amount of data extracted from the testing set.

\noindent
\textbf{Baseline methods:} \our\ is compared against the following six adversarial defense baselines.
\begin{itemize}[nosep,noitemsep,nolistsep]
    \item\textsc{DISP}~\citep{ZhouJCW19}: It is an embedding feature reconstruction method. It uses a perturbation discriminator to evaluate the probability that a token is perturbed and provides a set of potential perturbations. For each potential perturbation, an embedding estimator learns to restore the embedding of the original word based on the context.

    \item\textsc{FGWS}~\citep{MozesSKG21}: It uses frequency-guided word substitutions to exploit the frequency properties of adversarial word substitutions to detect adversaries.

    \item\textsc{RS\&V}~\citep{WangXH22}: It is a text reconstruction method based on the randomized substitution-to-vote strategy. \textsc{RS\&V} accumulates the logits of massive samples generated by randomly substituting the words in the adversaries with synonyms. 
\end{itemize}
Note that the rationale of choosing the above three baselines is their open source nature, while we can hardly reproduce the experimental results of other methods like \textsc{Textshield}~\citep{ShenZJC23}.


\section{Experimental Results}
\begin{table*}[hbtp]
		\centering
		\caption{The main adversarial detection and defense performance of \our\ on four public datasets. The victim model is \bert\ and the results in \textbf{bold} font indicate the best performance. We report the average accuracy of five random runs. The adversarial defense performance reported in previous works varies from adversarial attackers' implementations. For fair comparisons, all the baseline experiments are re-implemented based on the latest adversarial attackers from the Textattack library to avoid biases. ``\textsc{TF}'' indicates \textfooler.}
		\resizebox{\linewidth}{!}{
        \renewcommand{\arraystretch}{1.3}
            \begin{tabular}{c@{\hspace{0.1cm}}|c@{\hspace{0.1cm}}|c@{\hspace{0.02cm}}c@{\hspace{0.02cm}}c@{\hspace{0.02cm}}c@{\hspace{0.02cm}}c@{\hspace{0.02cm}}|c@{\hspace{0.02cm}}c@{\hspace{0.02cm}}c@{\hspace{0.02cm}}c@{\hspace{0.02cm}}c@{\hspace{0.02cm}}|c@{\hspace{0.02cm}}c@{\hspace{0.02cm}}c@{\hspace{0.02cm}}c@{\hspace{0.02cm}}c@{\hspace{0.02cm}}|c@{\hspace{0.02cm}}c@{\hspace{0.02cm}}c@{\hspace{0.02cm}}c@{\hspace{0.02cm}}c@{\hspace{0.02cm}}}
    \hline
    \multirow{2}[1]{*}{\textsc{Defender}} & \multirow{2}[1]{*}{\textsc{Attacker}} & \multicolumn{5}{c}{\agnews (4-category)}            & \multicolumn{5}{|c}{\yahoo (10-category)}            & \multicolumn{5}{|c}{\sst2 (2-category)}            & \multicolumn{5}{|c}{\amazon (2-category)} \\
\cline{3-22}          &       & \multicolumn{1}{c}{\textsc{NtA}} & \multicolumn{1}{c}{\textsc{AtA}} & \multicolumn{1}{c}{\textsc{DtA}} & \multicolumn{1}{c}{\textsc{DfA}} & \multicolumn{1}{c}{\textsc{RpA}} & \multicolumn{1}{|c}{\textsc{NtA}} & \multicolumn{1}{c}{\textsc{AtA}} & \multicolumn{1}{c}{\textsc{DtA}} & \multicolumn{1}{c}{\textsc{DtA}} & \multicolumn{1}{c}{\textsc{RpA}} & \multicolumn{1}{|c}{\textsc{NtA}} & \multicolumn{1}{c}{\textsc{AtA}} & \multicolumn{1}{c}{\textsc{DtA}} & \multicolumn{1}{c}{\textsc{DfA}} & \multicolumn{1}{c}{\textsc{RpA}} & \multicolumn{1}{|c}{\textsc{NtA}} & \multicolumn{1}{c}{\textsc{AtA}} & \multicolumn{1}{c}{\textsc{DtA}} & \multicolumn{1}{c}{\textsc{DfA}} & \multicolumn{1}{c}{\textsc{RpA}} \\
    \hline
    \hline
    \multirow{3}[2]{*}{\disp} 
          & \pwws  &       & $32.09$ & $55.49$ & $57.82$ & $68.23$ &       & $5.70$ & $61.67$ & $54.95$ & $50.24$ &       & $23.44$ & $38.93$ & $34.46$ & $35.33$ &       & $15.56$ & $41.90$ & $45.92$ & $59.80$ \\
          & \textsc{TF} & $94.13$ & $50.50$ & $53.78$ & $56.18$ & $70.16$ & $75.63$ & $13.60$ & $50.73$ & $57.48$ & $53.18$ & $91.24$ & $16.21$ & $37.80$ & $34.37$ & $37.16$ & $93.67$ & $21.77$ & $43.10$ & $47.15$ & $60.56$ \\
          & \bae   &       & $74.80$ & $45.26$ & $45.75$ & $81.39$ &       & $27.50$ & $54.82$ & $53.75$ & $50.90$ &       & $35.21$ & $36.59$ & $37.51$ & $42.22$ &       & $44.00$ & $40.28$ & $42.74$ & $61.85$ \\
    \hline
    \hline

    \multirow{3}[2]{*}{\fgws} 
          & \pwws  &       & $32.09$ & $65.24$ & $68.35$ & $71.78$ &       & $5.70$ & $65.83$ & $61.46$ & $53.28$ &       & $23.44$ & $40.28$ & $40.38$ & $39.20$ &       & $15.56$ & $44.47$ & $56.89$ & $60.29$ \\
          & \textsc{TF} & $94.25$ & $50.50$ & $68.88$ & $70.71$ & $73.40$ & $76.24$ & $13.60$ & $68.57$ & $65.17$ & $54.53$ & $91.34$ & $16.21$ & $42.79$ & $41.05$ & $41.53$ & $94.26$ & $21.77$ & $45.75$ & $58.74$ & $61.51$ \\
          & \bae   &       & $74.80$ & $44.29$ & $47.95$ & $83.57$ &       & $27.50$ & $58.63$ & $56.33$ & $52.94$ &       & $35.21$ & $43.83$ & $48.37$ & $44.90$ &       & $44.00$ & $42.26$ & $43.04$ & $64.63$ \\
    \hline
    \hline
    \multirow{3}[2]{*}{\rsv} 
          & \pwws  &       & $32.09$ & $83.67$ & $84.96$ & $83.80$ &       & $5.70$ & $65.01$ & $65.22$ & $57.22$ &       & $23.44$ & $36.90$ & $37.10$ & $38.54$ &       & $15.56$ & $29.60$ & $45.30$ & $46.17$ \\
          & \textsc{TF} & $94.14$ & $50.50$ & $82.44$ & $83.45$ & $82.53$ & $76.39$ & $13.60$ & $74.21$ & $74.54$ & $58.10$ & $91.55$ & $16.21$ & $39.70$ & $38.40$ & $39.70$ & $94.32$ & $21.77$ & $40.70$ & $42.30$ & $55.70$ \\
          & \bae   &       & $74.80$ & $46.98$ & $48.67$ & $86.90$ &       & $27.50$ & $37.41$ & $37.88$ & $62.27$ &       & $35.21$ & $19.84$ & $20.92$ & $43.65$ &       & $44.00$ & $38.59$ & $39.01$ & $65.03$ \\
    \hline
    \hline
    \multirow{3}[2]{*}{\our} 
          & \pwws  &       & $32.09$ & $\mathbf{90.11}$ & $\mathbf{95.88}$ & $\mathbf{92.36}$ &       & $5.70$ & $\mathbf{87.33}$ & $\mathbf{92.47}$ & $\mathbf{69.40}$ &       & $23.44$ & $\mathbf{94.03}$ & $\mathbf{98.62}$ & $\mathbf{89.85}$ &       & $15.56$ & $\mathbf{97.33}$ & $\mathbf{99.99}$ & $\mathbf{94.42}$ \\
          & \textsc{TF} & $94.30$ & $50.50$ & $\mathbf{90.29}$ & $\mathbf{96.76}$ & $\mathbf{92.14}$ & $76.45$ & $13.60$ & $\mathbf{87.49}$ & $\mathbf{93.54}$ & $\mathbf{70.50}$ & $91.70$ & $16.21$ & $\mathbf{94.03}$ & $\mathbf{99.86}$ & $\mathbf{89.72}$ & $94.24$ & $21.77$ & $\mathbf{93.85}$ & $\mathbf{99.99}$ & $\mathbf{93.96}$ \\
          & \bae   &       & $74.80$ & $\mathbf{57.55}$ & $\mathbf{96.25}$ & $\mathbf{93.64}$ &       & $27.50$ & $\mathbf{82.46}$ & $\mathbf{96.30}$ & $\mathbf{73.06}$ &       & $35.21$ & $\mathbf{78.99}$ & $\mathbf{99.28}$ & $\mathbf{89.77}$ &       & $44.00$ & $\mathbf{80.55}$ & $\mathbf{99.99}$ & $\mathbf{93.89}$ \\

    \hline
    \end{tabular}%
		}
		\label{tab:main}
    \vspace{-10pt}
\end{table*}

\subsection{Adversary detection performance}

Results shown in \pref{tab:main} demonstrate the effectiveness of the adversarial detector in \our. This in-victim-model adversarial detector, trained in conjunction with the standard classifier, accurately identifies adversaries across most datasets. Compared to the previous adversary detection-based defense~\citep{MozesSKG21,WangXH22,ShenZJC23}, the in-victim-model adversarial detector identifies the adversaries with no extra cost. On the other hand, our evaluation confirms a very low false positive rate ($\approx 2\%$) of adversary detection on natural examples, resulting in a very slight performance degradation on natural examples. Further, the adaptability of \our\ to previously unseen attack methods is evidenced in~\pref{tab:rq1}, highlighting the versatility of our adversarial detector. It excels at identifying adversaries by detecting disruptions introduced by malicious attackers, such as grammar errors and word misuse. Note that detection performance on the \agnews\ dataset is lower due to the absence of news data in the \textsc{BERT} training corpus, as discussed in Table $8$ of~\citet{HeGC21}.

\subsection{Adversary defense performance}
\label{sec:results_defense}

As for the adversary defense, \our\ outperforms existing methods across all datasets, as outlined in \pref{tab:main}. When we focus on correctly identified adversaries, \our\ can effectively repair up to $92\%$ to $99\%$ of them, even on the challenging 10-category \texttt{Yahoo} datasets. Our research also sheds light on the limitations of unsupervised text-level and feature-level reconstruction methods, as reported in studies such as~\citet{ZhouJCW19,MozesSKG21,WangXH22}. These methods struggle to rectify the deep semantics in adversaries, rendering them inefficient and inferior. Additionally, we find that previous methods are not robust when defending against adversaries in short texts, as evidenced by their failure on the \sst2\ and \amazon\ datasets. \our\ consistently achieves higher defense accuracy, particularly on binary classification datasets. In summary, \our\ employs adversarial attackers to repair adversaries' deep semantics and minimize edits in the text space, resulting in satisfactory adversarial defense. We emphasize the importance of dedicated deep semantics repair in the context of adversarial defense against unsupervised features and text space reconstruction.

\vspace{-.5em}
\begin{table}[t!]
        \centering
		\caption{The performance of \our\ \textbf{without pseudo-similarity filtering} (colored numbers indicate performance declines in the ablation). The metrics not unaffected by the pseudo-similarity filtering are omitted.}
		\resizebox{\linewidth}{!}{
                \begin{tabular}{c|c|c|c}
				\hline
				\textsc{Dataset} & \textsc{Attacker}   & \textsc{DtA}  & \textsc{RpA}  \\
				\hline
				\multirow{3}[1]{*}{\texttt{\agnews}}  & \pwws  & $94.19(\textcolor{red}{-1.69\downarrow})$  & $90.80(\textcolor{red}{-1.56\downarrow})$  \\
				& \textsc{TF}  & $94.26(\textcolor{red}{-2.50\downarrow})$ & $91.35(\textcolor{red}{-0.79\downarrow})$ \\
    		& \bae         & $92.98(\textcolor{red}{-3.27\downarrow})$ & $91.44(\textcolor{red}{-2.20\downarrow})$ \\
				\hline  
				\multirow{3}[1]{*}{\texttt{\yahoo}}   & \pwws  & $88.04(\textcolor{red}{-4.43\downarrow})$ & $65.38(\textcolor{red}{-4.02\downarrow})$ \\
				& \textsc{TF}  & $91.28(\textcolor{red}{-2.26\downarrow})$ & $67.48(\textcolor{red}{-3.02\downarrow})$ \\
        		& \bae         & $92.48(\textcolor{red}{-3.84\downarrow})$ & $71.35(\textcolor{red}{-1.71\downarrow})$ \\
    		\hline
		      \multirow{3}[1]{*}{\texttt{\sst2}}    & \pwws  & $98.12(\textcolor{red}{-0.50\downarrow})$ & $87.80(\textcolor{red}{-2.05\downarrow})$ \\
				& \textsc{TF}  & $98.03(\textcolor{red}{-1.83\downarrow})$ & $88.40(\textcolor{red}{-1.32\downarrow})$ \\
        		& \bae         & $95.87(\textcolor{red}{-3.41\downarrow})$ & $87.52(\textcolor{red}{-2.25\downarrow})$ \\
        		\hline
		      \multirow{3}[1]{*}{\texttt{\amazon}}  & \pwws  & $99.99(\textcolor{black}{~0.00})$ & $94.40(\textcolor{red}{-0.02\downarrow})$ \\
				& \textsc{TF}  & $98.92(\textcolor{red}{-1.07\downarrow})$ & $93.31(\textcolor{red}{-0.65\downarrow})$ \\
        		& \bae         & $98.53(\textcolor{red}{-1.41\downarrow})$ & $93.62(\textcolor{red}{-0.27\downarrow})$ \\
        		\hline
			\end{tabular}
		}
		\label{tab:ablation}
          \vspace{-15pt}
\end{table}

\subsection{Ablation experiment}

We conducted ablation experiments to assess the effectiveness of pseudo-semantic similarity filtering (\pref{sec:similarity}). It exclusively affects the defense process, so we have omitted the unaffected metrics, such as the detection accuracy in~\pref{tab:main}. From the results shown in~\pref{tab:ablation}, we find that the adversarial defense performance of \our\ without this filtering strategy is notably inferior ($\approx 1\%$) in most cases. Further, the degradation in defense performance is more pronounced in the case of the \agnews\ and \yahoo\ datasets compared to the \sst2\ and \amazon\ datasets. This discrepancy is attributed to the larger vocabularies and longer text lengths in the \agnews\ and \yahoo\ datasets, resulting in diversified repaired examples in terms of similarity.

\subsection{Further research questions}
\label{sec:rq}

We discuss more findings about \our~ by answering the following research questions (RQs).

\noindent
\textbf{RQ1: How is the generalization ability of \our\ to unknown attackers?}

\noindent
\underline{\textit{Methods}:} To assess the generalization ability of the in-victim-model adversarial detector in \our, we have conducted experiments among various state-of-the-art adversarial attackers: \pso, \iga, \wordbugger, and \clare, which were not included in the training of the adversarial detector in \our. Note that better adversarial detection and defense performance against unknown adversarial attackers indicates a superior generalizability of \our.

\begin{table}[ht]
	\centering
	\caption{Performance of \our\ for adversarial detection and defense \textbf{against unknown adversarial attacks}. }
	\resizebox{.95\linewidth}{!}{
        \renewcommand{\arraystretch}{1.05}
		\begin{tabular}{c|c|c|c|c|c}
			\hline
            \textsc{Dataset} & \textsc{Attacker} & \textsc{AtA} & \textsc{DtA} & \textsc{DfA}  & \textsc{RpA} \\
			\hline
		\multirow{3.5}[2]{*}{\agnews} & \pso   & $14.83$ & $68.46$ & $67.82$ & $90.39$ \\
			& \iga    & $26.87$ & $76.74$ & $74.59$ & $92.33$ \\
			& \wordbugger & $45.53$  & $72.73$ & $87.23$ & $89.33$ \\
			& \clare & $8.46$ & $62.78$ & $61.54$ & $64.78$ \\
			\hline
			\multirow{3.5}[2]{*}{\yahoo} & \pso   & $6.28$ & $80.26$ & $76.89$ & $87.82$ \\
			& \iga    & $14.75$ &  $82.69$ & $81.02$ & $54.55$ \\
			& \wordbugger &  $51.34$  & $72.73$ & $87.10$ & $62.27$ \\
			& \clare & $3.56$ & $64.85$ & $62.40$ & $52.47$ \\
			\hline
			\multirow{3.5}[2]{*}{\sst2}  & \pso   & $7.95$  & $87.50$ & $87.50$ & $82.61$ \\
			& \iga    & $18.39$  & $89.33$ & $98.67$ & $87.68$ \\
			& \wordbugger & $30.67$ & $95.44$ & $83.59$ & $81.90$ \\
			& \clare & $2.59$ & $62.50$ & $59.37$ & $65.30$ \\
			\hline
			\multirow{3.5}[2]{*}{\amazon} & \pso   & $5.76$  & $90.48$ & $90.48$ & $91.55$ \\
			& \iga    & $14.91$ & $92.31$ & $92.31$ & $94.65$ \\
			& \wordbugger & $43.43$ & $87.04$ & $85.19$ & $86.87$ \\
			& \clare & $3.25$ & $60.44$ & $59.37$ & $62.94$ \\
			\hline

		\end{tabular}%
	}
	\label{tab:rq1}%
    \vspace{-15pt}
\end{table}

\noindent
\underline{\textit{Results}:} From the results in~\pref{tab:rq1}, we find that \our\ can identify up to $98.67\%$ of adversaries on both the \sst2\ and \amazon\ datasets when considering adversarial detection performance. In terms of adversarial defense, \our\ is capable of repairing a substantial number of adversaries generated by various unknown attack methods (up to $87.68\%$ and $94.65\%$ on the \sst2\ and \amazon\ datasets, respectively). However, \our\ experiences a decline in performance in identifying and defending against adversaries when facing the challenging \clare\ attack. This performance degradation is likely attributed to their ineffective adversarial detection, which could potentially be improved by training \clare-based adversaries for adversarial detection within \our. In summary, \our\ has demonstrated robust generalization ability, effectively detecting and repairing a wide array of adversaries generated by unknown attackers.

\noindent
\textbf{RQ2: Does perturbation defocusing really repair adversaries?} 

\begin{figure}[ht]
	\centering
	\includegraphics[width=\linewidth]{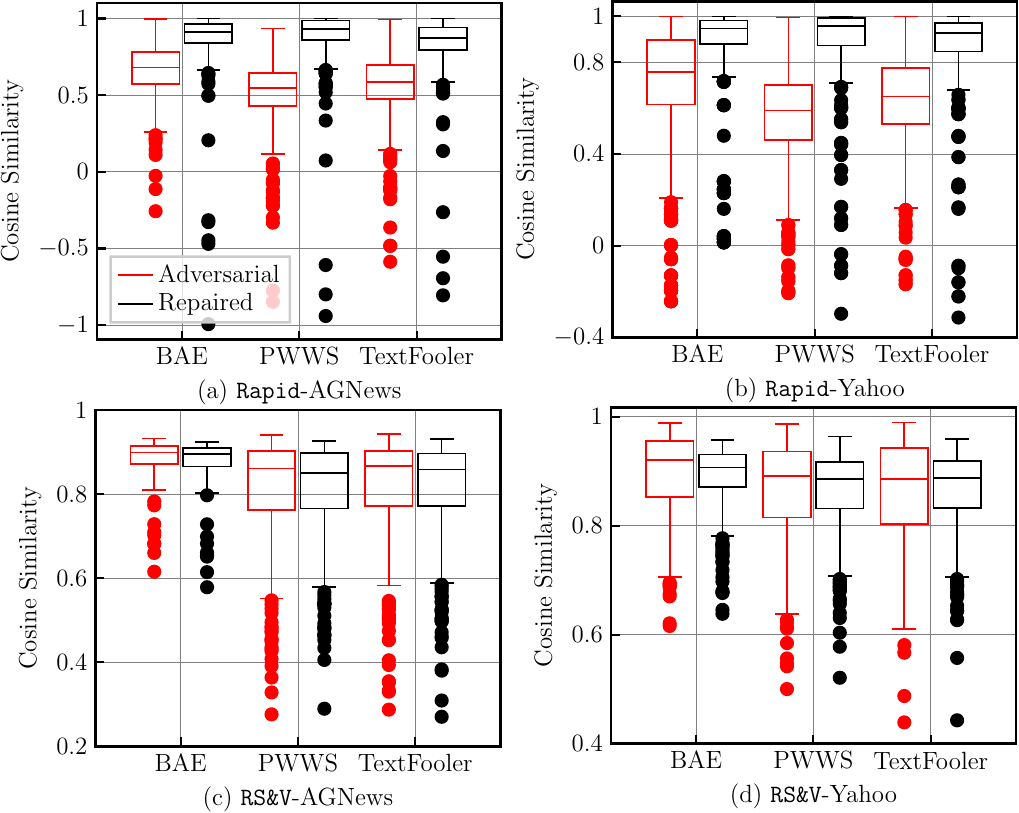}
	\caption{Box plots of semantic cosine similarity score distributions on multi-categorial datasets. Similar to \pref{fig:intro}, \our\ is more competent to repair semantics according to the feature similarity score distributions.}
	\label{fig:rq2-cosine}
    \vspace{-10pt}
\end{figure}

\noindent
\underline{\textit{Methods}:} To address this RQ, we investigate the discrepancy between adversaries and their repaired counterparts in the feature space. Specifically, we employ three attackers (i.e., \bae, \pwws, \textfooler) to generate adversaries and their corresponding repaired examples, considering a random selection of $1,000$ natural examples. Using the victim model, we encode these examples into the feature space and evaluate the cosine similarity between adversary-natural example pairs and repaired adversary-natural example pairs. The larger cosine similarity scores indicate better performance in repairing the deep semantics in the adversaries.

\noindent
\underline{\textit{Results}:} The box plots in Figures~\ref{fig:intro} and \ref{fig:rq2-cosine} show the similarity score distributions collected from pairwise semantic similarity assessments. The semantic similarity score distributions (e.g., the median similarity scores of repaired examples are always larger than the adversaries) from these plots reveal a notable global similarity between the natural examples and repaired examples by \our, which means \our\ does repair the deep semantics of the adversaries. Conversely, it is apparent that the similarity scores of the repaired examples obtained using \rsv\ are indistinguishable from the adversarial examples across all datasets. This situation happens to many of the existing adversarial defense methods. In conclusion, our observations show the ability of \our\ to effectively repair the deep semantics of adversaries.



\noindent
\textbf{RQ3: How does the inherent robustness of the victim model affect \our?}

\begin{table}[t!]
		\centering
		\caption{The performance of \our\ on four public datasets \textbf{based on the victim model \deberta}. The numbers in \textcolor{red}{\underline{red}} color indicate performance declines compared to the \bert-based \our.}
		\resizebox{\linewidth}{!}{
        \renewcommand{\arraystretch}{1.2}
  	\begin{tabular}{c|c|c|c|c|c|c}
				\hline
                \textsc{Dataset} & \textsc{Attacker}  & \textsc{NtA} & \textsc{AtA} & \textsc{DtA}   & \textsc{DfA}  & \textsc{RpA}  \\
				\hline
				\multirow{3}[1]{*}{\agnews}  & \pwws  & \multirow{3}[1]{*}{$96.69$} & $62.77$ & $96.47$ & $98.47$ & $93.12$ \\
				& \textsc{TF} &   & $39.85$  & $91.41$ & $\textcolor{red}{\underline{95.90}\downarrow}$ & $93.69$ \\ 
    		& \bae         &   & $81.64$  & $90.20$ & $97.92$ & $\textcolor{red}{\underline{93.40}\downarrow}$ \\
				\hline
				\multirow{3}[1]{*}{\yahoo}   &  \pwws  & \multirow{3}[1]{*}{$78.63$} & $15.70$ & $88.91$ & $92.64$ & $70.47$ \\
				& \textsc{TF}  &  & ~~$6.19$ & $89.32$ & $92.60$ & $\textcolor{red}{\underline{69.96}\downarrow}$	\\
        		& \bae         &  & $47.50$ & $90.25$ & $\textcolor{red}{\underline{93.74}\downarrow}$ & $\textcolor{red}{\underline{72.12}\downarrow}$ \\
				\hline
		      \multirow{3}[1]{*}{\sst2}     & \pwws    & \multirow{3}[1]{*}{$95.01$} & $37.14$ & $95.21$ & $98.42$ & $94.15$ \\
				& \textsc{TF}  &   & $22.59$ & $\textcolor{red}{\underline{93.06}\downarrow}$ & $99.08$ & $94.58$ \\
        		& \bae         &   & $38.84$ & $80.82$ & $98.59$ & $94.16$ \\
        		\hline
		      \multirow{3}[1]{*}{\amazon}   & \pwws & \multirow{3}[1]{*}{$95.51$} & $22.72$ & $97.62$ & $99.99$ & $94.55$ \\
				& \textsc{TF}  &   & $23.95$ & $94.91$ & $99.99$ & $94.84$ \\
        		& \bae         &   & $56.65$ & $82.71$ & $99.99$ & $94.50$ \\
        		\hline
			\end{tabular}
		}
		\label{tab:new_rq3}
          \vspace{-15pt}
\end{table}

\noindent
\underline{\textit{Methods}:} We assessed the impact of the inherent robustness of the victim model, focusing on \deberta, a cutting-edge PLM utilized across various tasks. Specifically, we trained a victim model based on \deberta, replicating the experimental setup and evaluating the performance variation of \our\ based on this \deberta\ victim model.

\noindent
\underline{\textit{Results}:} As in~\pref{tab:new_rq3}, the \deberta-based victim model demonstrates superior accuracy under adversarial attacks, indicating higher inherent robustness in \deberta\ compared to the victim model built on \bert. In particular, \deberta-based \our\ excels in identifying adversaries across all classification datasets, especially on the binary datasets. The performance in adversarial detection and defense follows a similar upward trajectory. Emphasizing the substantial influence of the victim model's robustness on our method, particularly in enhancing adversarial detection and defense.




\section{Related Works}
\label{sec:related_works}

Prior research on adversarial defense can be classified into three categories: adversarial training-based methods~\citep{MiyatoDG17,ZhuCGSGL20,IvgiB21}; context reconstruction-based methods \citep{PruthiDL19,LiuLYLSLS20,MozesSKG21,KellerME21,ChenFZCH21,XuZZLHCH22,LiSZMQ22,SwenorK22}; and feature reconstruction-based methods\citep{ZhouJCW19,JonesJRL20,WangH0021}. Some studies~\citep{Wang0D021} also investigated hybrid defense methods. 
As for the adversarial training-based methods, they are notorious for the performance degradation of natural examples. They can improve the robustness of PLMs by fine-tuning, yet increasing the cost of model training caused by catastrophic forgetting~\citep{DongLLYZ21}. Text reconstruction-based methods, such as word substitution~~\citep{MozesSKG21,BaoWZ21} and translation-based reconstruction, may fail to identify semantically repaired adversaries or introduce new malicious perturbations~\citep{SwenorK22}. Feature reconstruction methods, on the other hand, may struggle to repair typo attacks~~\citep{LiuZWLC20,TanJVK20,JonesJRL20}, sentence-level attacks~\citep{ZhaoDS18,ChengJM19}, and other unknown attacks. 
There are some works towards the adversarial detection and defense joint task\citep{ZhouJCW19,MozesSKG21,WangXH22}. However, these adversarial detection methods may be ineffective for unknown adversarial attackers and can hardly alleviate resource waste in adversarial defense. Another similar work to \our\ is \texttt{Textshield}~\citep{ShenZJC23}, which aims to defend against word-level adversarial attacks by detecting adversarial sentences based on a saliency-based detector and fixing the adversarial examples using a corrector. 
Overall, our study focuses on maintaining the semantics by introducing minimal safe perturbations into adversaries, thus alleviating the semantic shifting problem in all reconstruction-based works.


\section{Conclusion}
\label{sec:conclusion}

We propose a novel adversarial defense method, i.e., perturbation defocusing, to repair semantics in adversarial examples. \our\ addresses the semantic shifting problem in the previous studies. \our\ shows an outstanding performance in repairing adversarial examples (up to $\approx 99\%$ of correctly identified adversarial examples). It is believed that perturbation defocusing has the potential to significantly shift the landscape of textual adversarial defense.

\section*{Limitations}
One limitation of the proposed method is that it tends to introduce new perturbations into the adversaries, which may lead to semantic shifts. This may be unsafe for some tasks, e.g., machine translation. Furthermore, the method requires a large amount of computational resources to generate the adversaries during the training phase, which may be a limitation in some scenarios. Finally, the method has not been tested on a wide range of NLP tasks and domains, and further evaluations on other tasks and domains are necessary to fully assess its capabilities.

\bibliography{custom}

\begin{thebibliography}{54}
\expandafter\ifx\csname natexlab\endcsname\relax\def\natexlab#1{#1}\fi

\bibitem[{Alzantot et~al.(2018)Alzantot, Sharma, Elgohary, Ho, Srivastava, and
  Chang}]{AlzantotSEHSC18}
Moustafa Alzantot, Yash Sharma, Ahmed Elgohary, Bo{-}Jhang Ho, Mani~B.
  Srivastava, and Kai{-}Wei Chang. 2018.
\newblock \href {https://doi.org/10.18653/v1/d18-1316} {Generating natural
  language adversarial examples}.
\newblock In \emph{EMNLP'18: Proc. of the 2018 Conference on Empirical Methods
  in Natural Language Processing}, pages 2890--2896. Association for
  Computational Linguistics.

\bibitem[{Bao et~al.(2021)Bao, Wang, and Zhao}]{BaoWZ21}
Rongzhou Bao, Jiayi Wang, and Hai Zhao. 2021.
\newblock \href {https://doi.org/10.18653/v1/2021.findings-acl.287} {Defending
  pre-trained language models from adversarial word substitution without
  performance sacrifice}.
\newblock In \emph{ACL-IJCNLP'21: Findings of the 2021 Conference of the
  Association for Computational Linguistics: {ACL-IJCNLP} 2021, Online Event,
  August 1-6, 2021}, volume {ACL-IJCNLP} 2021 of \emph{Findings of {ACL}},
  pages 3248--3258. Association for Computational Linguistics.

\bibitem[{Boucher et~al.(2022)Boucher, Shumailov, Anderson, and
  Papernot}]{BoucherSAP22}
Nicholas Boucher, Ilia Shumailov, Ross Anderson, and Nicolas Papernot. 2022.
\newblock \href {https://doi.org/10.1109/SP46214.2022.9833641} {Bad characters:
  Imperceptible {NLP} attacks}.
\newblock In \emph{43rd {IEEE} Symposium on Security and Privacy, {SP} 2022,
  San Francisco, CA, USA, May 22-26, 2022}, pages 1987--2004. {IEEE}.

\bibitem[{Chen et~al.(2021)Chen, Fan, Zhang, Chen, and Huang}]{ChenFZCH21}
Guandan Chen, Kai Fan, Kaibo Zhang, Boxing Chen, and Zhongqiang Huang. 2021.
\newblock \href {https://doi.org/10.18653/v1/2021.findings-acl.281} {Manifold
  adversarial augmentation for neural machine translation}.
\newblock In \emph{ACL-IJCNLP'21: Findings of the 2021 Conference of the
  Association for Computational Linguistics}, pages 3184--3189. Association for
  Computational Linguistics.

\bibitem[{Cheng et~al.(2019)Cheng, Jiang, and Macherey}]{ChengJM19}
Yong Cheng, Lu~Jiang, and Wolfgang Macherey. 2019.
\newblock \href {https://doi.org/10.18653/v1/p19-1425} {Robust neural machine
  translation with doubly adversarial inputs}.
\newblock In \emph{ACL'19: Proc. of the 57th Conference of the Association for
  Computational Linguistics}, pages 4324--4333. Association for Computational
  Linguistics.

\bibitem[{Devlin et~al.(2019)Devlin, Chang, Lee, and Toutanova}]{DevlinCLT19}
Jacob Devlin, Ming{-}Wei Chang, Kenton Lee, and Kristina Toutanova. 2019.
\newblock \href {https://doi.org/10.18653/v1/n19-1423} {{BERT:} pre-training of
  deep bidirectional transformers for language understanding}.
\newblock In \emph{NAACL-HLT'19: Proc. of the 2019 Conference of the North
  American Chapter of the Association for Computational Linguistics}, pages
  4171--4186. Association for Computational Linguistics.

\bibitem[{Dong et~al.(2021{\natexlab{a}})Dong, Luu, Ji, and Liu}]{DongLJ021}
Xinshuai Dong, Anh~Tuan Luu, Rongrong Ji, and Hong Liu. 2021{\natexlab{a}}.
\newblock \href {https://openreview.net/forum?id=ks5nebunVn\_} {Towards
  robustness against natural language word substitutions}.
\newblock In \emph{ICLR'21: Proc. of the 9th International Conference on
  Learning Representations}. OpenReview.net.

\bibitem[{Dong et~al.(2021{\natexlab{b}})Dong, Luu, Lin, Yan, and
  Zhang}]{DongLLYZ21}
Xinshuai Dong, Anh~Tuan Luu, Min Lin, Shuicheng Yan, and Hanwang Zhang.
  2021{\natexlab{b}}.
\newblock \href
  {https://proceedings.neurips.cc/paper/2021/hash/22b1f2e0983160db6f7bb9f62f4dbb39-Abstract.html}
  {How should pre-trained language models be fine-tuned towards adversarial
  robustness?}
\newblock In \emph{NeurIPS'21: Proc. of the 2021 Conference on Neural
  Information Processing Systems}, pages 4356--4369.

\bibitem[{Ebrahimi et~al.(2018)Ebrahimi, Rao, Lowd, and Dou}]{EbrahimiRLD18}
Javid Ebrahimi, Anyi Rao, Daniel Lowd, and Dejing Dou. 2018.
\newblock \href {https://doi.org/10.18653/V1/P18-2006} {Hotflip: White-box
  adversarial examples for text classification}.
\newblock In \emph{Proceedings of the 56th Annual Meeting of the Association
  for Computational Linguistics, {ACL} 2018, Melbourne, Australia, July 15-20,
  2018, Volume 2: Short Papers}, pages 31--36. Association for Computational
  Linguistics.

\bibitem[{Gao et~al.(2018)Gao, Lanchantin, Soffa, and Qi}]{GaoLSQ18}
Ji~Gao, Jack Lanchantin, Mary~Lou Soffa, and Yanjun Qi. 2018.
\newblock \href {https://doi.org/10.1109/SPW.2018.00016} {Black-box generation
  of adversarial text sequences to evade deep learning classifiers}.
\newblock In \emph{SP'18: Proc. of the 2018 {IEEE} Security and Privacy
  Workshops}, pages 50--56. {IEEE} Computer Society.

\bibitem[{Garg and Ramakrishnan(2020)}]{GargR20}
Siddhant Garg and Goutham Ramakrishnan. 2020.
\newblock \href {https://doi.org/10.18653/v1/2020.emnlp-main.498} {{BAE:}
  bert-based adversarial examples for text classification}.
\newblock In \emph{EMNLP'20: Proc. of the 2020 Conference on Empirical Methods
  in Natural Language Processing}, pages 6174--6181. Association for
  Computational Linguistics.

\bibitem[{Gowal et~al.(2021)Gowal, Rebuffi, Wiles, Stimberg, Calian, and
  Mann}]{GowalRWSCM21}
Sven Gowal, Sylvestre{-}Alvise Rebuffi, Olivia Wiles, Florian Stimberg,
  Dan~Andrei Calian, and Timothy~A. Mann. 2021.
\newblock \href
  {https://proceedings.neurips.cc/paper/2021/hash/21ca6d0cf2f25c4dbb35d8dc0b679c3f-Abstract.html}
  {Improving robustness using generated data}.
\newblock In \emph{NeurIPS'21: Advances in Neural Information Processing
  Systems}, pages 4218--4233.

\bibitem[{Guo et~al.(2021)Guo, Sablayrolles, J{\'{e}}gou, and Kiela}]{GuoSJK21}
Chuan Guo, Alexandre Sablayrolles, Herv{\'{e}} J{\'{e}}gou, and Douwe Kiela.
  2021.
\newblock \href {https://doi.org/10.18653/V1/2021.EMNLP-MAIN.464}
  {Gradient-based adversarial attacks against text transformers}.
\newblock In \emph{Proceedings of the 2021 Conference on Empirical Methods in
  Natural Language Processing, {EMNLP} 2021, Virtual Event / Punta Cana,
  Dominican Republic, 7-11 November, 2021}, pages 5747--5757. Association for
  Computational Linguistics.

\bibitem[{He et~al.(2021)He, Gao, and Chen}]{HeGC21}
Pengcheng He, Jianfeng Gao, and Weizhu Chen. 2021.
\newblock \href {http://arxiv.org/abs/2111.09543} {Debertav3: Improving deberta
  using electra-style pre-training with gradient-disentangled embedding
  sharing}.
\newblock \emph{CoRR}, abs/2111.09543.

\bibitem[{Ivgi and Berant(2021)}]{IvgiB21}
Maor Ivgi and Jonathan Berant. 2021.
\newblock \href {https://doi.org/10.18653/v1/2021.emnlp-main.115} {Achieving
  model robustness through discrete adversarial training}.
\newblock In \emph{EMNLP'21: Proc. of the 2021 Conference on Empirical Methods
  in Natural Language Processing}, pages 1529--1544. Association for
  Computational Linguistics.

\bibitem[{Jin et~al.(2020)Jin, Jin, Zhou, and Szolovits}]{JinJZS20}
Di~Jin, Zhijing Jin, Joey~Tianyi Zhou, and Peter Szolovits. 2020.
\newblock \href {https://ojs.aaai.org/index.php/AAAI/article/view/6311} {Is
  {BERT} really robust? {A} strong baseline for natural language attack on text
  classification and entailment}.
\newblock In \emph{AAAI'20: Proc. of the 34th {AAAI} Conference on Artificial
  Intelligence}, pages 8018--8025. {AAAI} Press.

\bibitem[{Jones et~al.(2020)Jones, Jia, Raghunathan, and Liang}]{JonesJRL20}
Erik Jones, Robin Jia, Aditi Raghunathan, and Percy Liang. 2020.
\newblock \href {https://doi.org/10.18653/v1/2020.acl-main.245} {Robust
  encodings: {A} framework for combating adversarial typos}.
\newblock In \emph{ACL'20: Proc. of the 58th Annual Meeting of the Association
  for Computational Linguistics Conference}, pages 2752--2765. Association for
  Computational Linguistics.

\bibitem[{Keller et~al.(2021)Keller, Mackensen, and Eger}]{KellerME21}
Yannik Keller, Jan Mackensen, and Steffen Eger. 2021.
\newblock \href {https://doi.org/10.18653/v1/2021.findings-acl.141}
  {Bert-defense: {A} probabilistic model based on {BERT} to combat cognitively
  inspired orthographic adversarial attacks}.
\newblock In \emph{ACL-IJCNLP'21: Findings of the 2021 Conference of the
  Association for Computational Linguistics}, volume {ACL-IJCNLP} 2021 of
  \emph{Findings of {ACL}}, pages 1616--1629. Association for Computational
  Linguistics.

\bibitem[{Li et~al.(2021)Li, Zhang, Peng, Chen, Brockett, Sun, and
  Dolan}]{LiZPCBSD21}
Dianqi Li, Yizhe Zhang, Hao Peng, Liqun Chen, Chris Brockett, Ming{-}Ting Sun,
  and Bill Dolan. 2021.
\newblock \href {https://doi.org/10.18653/v1/2021.naacl-main.400}
  {Contextualized perturbation for textual adversarial attack}.
\newblock In \emph{NAACL-HLT'21: Proc. of the 2021 Conference of the North
  American Chapter of the Association for Computational Linguistics}, pages
  5053--5069. Association for Computational Linguistics.

\bibitem[{Li et~al.(2019)Li, Ji, Du, Li, and Wang}]{LiJDLW19}
Jinfeng Li, Shouling Ji, Tianyu Du, Bo~Li, and Ting Wang. 2019.
\newblock \href
  {https://www.ndss-symposium.org/ndss-paper/textbugger-generating-adversarial-text-against-real-world-applications/}
  {Textbugger: Generating adversarial text against real-world applications}.
\newblock In \emph{26th Annual Network and Distributed System Security
  Symposium, {NDSS} 2019, San Diego, California, USA, February 24-27, 2019}.
  The Internet Society.

\bibitem[{Li et~al.(2020)Li, Ma, Guo, Xue, and Qiu}]{LiMGXQ20}
Linyang Li, Ruotian Ma, Qipeng Guo, Xiangyang Xue, and Xipeng Qiu. 2020.
\newblock \href {https://doi.org/10.18653/v1/2020.emnlp-main.500}
  {{BERT-ATTACK:} adversarial attack against {BERT} using {BERT}}.
\newblock In \emph{EMNLP'20: Proc. of the 2020 Conference on Empirical Methods
  in Natural Language Processing}, pages 6193--6202. Association for
  Computational Linguistics.

\bibitem[{Li et~al.(2022)Li, Song, Zeng, Ma, and Qiu}]{LiSZMQ22}
Linyang Li, Demin Song, Jiehang Zeng, Ruotian Ma, and Xipeng Qiu. 2022.
\newblock \href {https://doi.org/10.48550/arXiv.2203.14207} {Rebuild and
  ensemble: Exploring defense against text adversaries}.
\newblock \emph{CoRR}, abs/2203.14207.

\bibitem[{Liu et~al.(2020{\natexlab{a}})Liu, Zhang, Wang, Lin, and
  Chen}]{LiuZWLC20}
Hui Liu, Yongzheng Zhang, Yipeng Wang, Zheng Lin, and Yige Chen.
  2020{\natexlab{a}}.
\newblock \href {https://ojs.aaai.org/index.php/AAAI/article/view/6356} {Joint
  character-level word embedding and adversarial stability training to defend
  adversarial text}.
\newblock In \emph{AAAI'20: Proc. of the 34th {AAAI} Conference on Artificial
  Intelligence}, pages 8384--8391. {AAAI} Press.

\bibitem[{Liu et~al.(2020{\natexlab{b}})Liu, Liu, Yang, Liu, Su, Li, and
  She}]{LiuLYLSLS20}
Kai Liu, Xin Liu, An~Yang, Jing Liu, Jinsong Su, Sujian Li, and Qiaoqiao She.
  2020{\natexlab{b}}.
\newblock \href {https://ojs.aaai.org/index.php/AAAI/article/view/6357} {A
  robust adversarial training approach to machine reading comprehension}.
\newblock In \emph{AAAI'20: Proc. of the Thirty-Fourth {AAAI} Conference on
  Artificial Intelligence}, pages 8392--8400. {AAAI} Press.

\bibitem[{Miyato et~al.(2017)Miyato, Dai, and Goodfellow}]{MiyatoDG17}
Takeru Miyato, Andrew~M. Dai, and Ian~J. Goodfellow. 2017.
\newblock \href {https://openreview.net/forum?id=r1X3g2\_xl} {Adversarial
  training methods for semi-supervised text classification}.
\newblock In \emph{ICLR'17: Proc. of the 5th International Conference on
  Learning Representations}. OpenReview.net.

\bibitem[{Morris et~al.(2020)Morris, Lifland, Yoo, Grigsby, Jin, and
  Qi}]{MorrisLYGJQ20}
John~X. Morris, Eli Lifland, Jin~Yong Yoo, Jake Grigsby, Di~Jin, and Yanjun Qi.
  2020.
\newblock \href {https://doi.org/10.18653/V1/2020.EMNLP-DEMOS.16} {Textattack:
  {A} framework for adversarial attacks, data augmentation, and adversarial
  training in {NLP}}.
\newblock In \emph{Proceedings of the 2020 Conference on Empirical Methods in
  Natural Language Processing: System Demonstrations, {EMNLP} 2020 - Demos,
  Online, November 16-20, 2020}, pages 119--126. Association for Computational
  Linguistics.

\bibitem[{Mozes et~al.(2021)Mozes, Stenetorp, Kleinberg, and
  Griffin}]{MozesSKG21}
Maximilian Mozes, Pontus Stenetorp, Bennett Kleinberg, and Lewis~D. Griffin.
  2021.
\newblock \href {https://doi.org/10.18653/v1/2021.eacl-main.13}
  {Frequency-guided word substitutions for detecting textual adversarial
  examples}.
\newblock In \emph{EACL'21: Proc. of the 16th Conference of the European
  Chapter of the Association for Computational Linguistics}, pages 171--186.
  Association for Computational Linguistics.

\bibitem[{OpenAI(2023)}]{OpenAI23}
OpenAI. 2023.
\newblock \href {https://doi.org/10.48550/arXiv.2303.08774} {{GPT-4} technical
  report}.
\newblock \emph{CoRR}, abs/2303.08774.

\bibitem[{Pruthi et~al.(2019)Pruthi, Dhingra, and Lipton}]{PruthiDL19}
Danish Pruthi, Bhuwan Dhingra, and Zachary~C. Lipton. 2019.
\newblock \href {https://doi.org/10.18653/v1/p19-1561} {Combating adversarial
  misspellings with robust word recognition}.
\newblock In \emph{ACL'19: Proc. of the 57th Conference of the Association for
  Computational Linguistics}, pages 5582--5591. Association for Computational
  Linguistics.

\bibitem[{Ren et~al.(2019)Ren, Deng, He, and Che}]{RenDHC19}
Shuhuai Ren, Yihe Deng, Kun He, and Wanxiang Che. 2019.
\newblock \href {https://doi.org/10.18653/v1/p19-1103} {Generating natural
  language adversarial examples through probability weighted word saliency}.
\newblock In \emph{ACL'19: Proc. of the 57th Conference of the Association for
  Computational Linguistics}, pages 1085--1097. Association for Computational
  Linguistics.

\bibitem[{Rony et~al.(2019)Rony, Hafemann, Oliveira, Ayed, Sabourin, and
  Granger}]{RonyHOASG19}
J{\'{e}}r{\^{o}}me Rony, Luiz~G. Hafemann, Luiz~S. Oliveira, Ismail~Ben Ayed,
  Robert Sabourin, and Eric Granger. 2019.
\newblock \href {https://doi.org/10.1109/CVPR.2019.00445} {Decoupling direction
  and norm for efficient gradient-based {L2} adversarial attacks and defenses}.
\newblock In \emph{CVPR'19: {IEEE} Conference on Computer Vision and Pattern
  Recognition}, pages 4322--4330. Computer Vision Foundation / {IEEE}.

\bibitem[{Shen et~al.(2023)Shen, Zhang, Jiang, and Chen}]{ShenZJC23}
Lingfeng Shen, Ze~Zhang, Haiyun Jiang, and Ying Chen. 2023.
\newblock \href {https://openreview.net/pdf?id=xIWfWvKM7aQ} {Textshield: Beyond
  successfully detecting adversarial sentences in text classification}.
\newblock In \emph{ICLR'23: The Eleventh International Conference on Learning
  Representations}. OpenReview.net.

\bibitem[{Socher et~al.(2013)Socher, Perelygin, Wu, Chuang, Manning, Ng, and
  Potts}]{SocherPWCMNP13}
Richard Socher, Alex Perelygin, Jean Wu, Jason Chuang, Christopher~D. Manning,
  Andrew~Y. Ng, and Christopher Potts. 2013.
\newblock \href {https://aclanthology.org/D13-1170/} {Recursive deep models for
  semantic compositionality over a sentiment treebank}.
\newblock In \emph{Proceedings of the 2013 Conference on Empirical Methods in
  Natural Language Processing, {EMNLP} 2013, 18-21 October 2013, Grand Hyatt
  Seattle, Seattle, Washington, USA, {A} meeting of SIGDAT, a Special Interest
  Group of the {ACL}}, pages 1631--1642. {ACL}.

\bibitem[{Swenor and Kalita(2022)}]{SwenorK22}
Abigail Swenor and Jugal Kalita. 2022.
\newblock \href {http://arxiv.org/abs/2202.05758} {Using random perturbations
  to mitigate adversarial attacks on sentiment analysis models}.
\newblock \emph{CoRR}, abs/2202.05758.

\bibitem[{Tan et~al.(2020)Tan, Joty, Varshney, and Kan}]{TanJVK20}
Samson Tan, Shafiq~R. Joty, Lav~R. Varshney, and Min{-}Yen Kan. 2020.
\newblock \href {https://doi.org/10.18653/v1/2020.emnlp-main.455} {Mind your
  inflections! improving {NLP} for non-standard englishes with base-inflection
  encoding}.
\newblock In \emph{EMNLP'20: Proc. of the 2020 Conference on Empirical Methods
  in Natural Language Processing}, pages 5647--5663. Association for
  Computational Linguistics.

\bibitem[{Wang et~al.(2019{\natexlab{a}})Wang, Pruksachatkun, Nangia, Singh,
  Michael, Hill, Levy, and Bowman}]{WangPNSMHLB19}
Alex Wang, Yada Pruksachatkun, Nikita Nangia, Amanpreet Singh, Julian Michael,
  Felix Hill, Omer Levy, and Samuel~R. Bowman. 2019{\natexlab{a}}.
\newblock \href
  {https://proceedings.neurips.cc/paper/2019/hash/4496bf24afe7fab6f046bf4923da8de6-Abstract.html}
  {Superglue: {A} stickier benchmark for general-purpose language understanding
  systems}.
\newblock In \emph{NeurIPS'19: Advances in Neural Information Processing
  Systems}, pages 3261--3275.

\bibitem[{Wang et~al.(2019{\natexlab{b}})Wang, Singh, Michael, Hill, Levy, and
  Bowman}]{WangSMHLB19}
Alex Wang, Amanpreet Singh, Julian Michael, Felix Hill, Omer Levy, and
  Samuel~R. Bowman. 2019{\natexlab{b}}.
\newblock \href {https://openreview.net/forum?id=rJ4km2R5t7} {{GLUE:} {A}
  multi-task benchmark and analysis platform for natural language
  understanding}.
\newblock In \emph{ICLR'19: 7th International Conference on Learning
  Representations}. OpenReview.net.

\bibitem[{Wang et~al.(2020)Wang, Pei, Pan, Chen, Wang, and Li}]{WangPPCWL20}
Boxin Wang, Hengzhi Pei, Boyuan Pan, Qian Chen, Shuohang Wang, and Bo~Li. 2020.
\newblock \href {https://doi.org/10.18653/V1/2020.EMNLP-MAIN.495} {{T3:}
  tree-autoencoder constrained adversarial text generation for targeted
  attack}.
\newblock In \emph{Proceedings of the 2020 Conference on Empirical Methods in
  Natural Language Processing, {EMNLP} 2020, Online, November 16-20, 2020},
  pages 6134--6150. Association for Computational Linguistics.

\bibitem[{Wang et~al.(2022{\natexlab{a}})Wang, Xu, Liu, Cheng, and
  Li}]{WangXLCL22}
Boxin Wang, Chejian Xu, Xiangyu Liu, Yu~Cheng, and Bo~Li. 2022{\natexlab{a}}.
\newblock \href {https://doi.org/10.18653/V1/2022.FINDINGS-NAACL.14}
  {Semattack: Natural textual attacks via different semantic spaces}.
\newblock In \emph{Findings of the Association for Computational Linguistics:
  {NAACL} 2022, Seattle, WA, United States, July 10-15, 2022}, pages 176--205.
  Association for Computational Linguistics.

\bibitem[{Wang et~al.(2021{\natexlab{a}})Wang, Hao, Yang, and He}]{WangH0021}
Xiaosen Wang, Jin Hao, Yichen Yang, and Kun He. 2021{\natexlab{a}}.
\newblock \href {https://proceedings.mlr.press/v161/wang21a.html} {Natural
  language adversarial defense through synonym encoding}.
\newblock In \emph{UAI'21: Proc. of the 37th Conference on Uncertainty in
  Artificial Intelligence}, volume 161 of \emph{Proceedings of Machine Learning
  Research}, pages 823--833. {AUAI} Press.

\bibitem[{Wang et~al.(2022{\natexlab{b}})Wang, Xiong, and He}]{WangXH22}
Xiaosen Wang, Yifeng Xiong, and Kun He. 2022{\natexlab{b}}.
\newblock Detecting textual adversarial examples through randomized
  substitution and vote.
\newblock In \emph{{UAI}}, volume 180 of \emph{Proceedings of Machine Learning
  Research}, pages 2056--2065. {PMLR}.

\bibitem[{Wang et~al.(2021{\natexlab{b}})Wang, Yang, Deng, and He}]{Wang0D021}
Xiaosen Wang, Yichen Yang, Yihe Deng, and Kun He. 2021{\natexlab{b}}.
\newblock \href {https://ojs.aaai.org/index.php/AAAI/article/view/17648}
  {Adversarial training with fast gradient projection method against synonym
  substitution based text attacks}.
\newblock In \emph{AAAI'21: Proc. of the 35th {AAAI} Conference on Artificial
  Intelligence}, pages 13997--14005. {AAAI} Press.

\bibitem[{Wang et~al.(2023)Wang, Pang, Du, Lin, Liu, and Yan}]{WangPDLLY23}
Zekai Wang, Tianyu Pang, Chao Du, Min Lin, Weiwei Liu, and Shuicheng Yan. 2023.
\newblock \href {https://doi.org/10.48550/arXiv.2302.04638} {Better diffusion
  models further improve adversarial training}.
\newblock \emph{CoRR}, abs/2302.04638.

\bibitem[{Xu et~al.(2022)Xu, Zhang, Zheng, Li, Hsieh, Chang, and
  Huang}]{XuZZLHCH22}
Jianhan Xu, Cenyuan Zhang, Xiaoqing Zheng, Linyang Li, Cho{-}Jui Hsieh,
  Kai{-}Wei Chang, and Xuanjing Huang. 2022.
\newblock \href {https://doi.org/10.18653/v1/2022.findings-acl.134} {Towards
  adversarially robust text classifiers by learning to reweight clean
  examples}.
\newblock In \emph{ACL'22: Findings of the 2022 Conference of the Association
  for Computational Linguistics}, pages 1694--1707. Association for
  Computational Linguistics.

\bibitem[{Xu et~al.(2023)Xu, Sun, Goldblum, Goldstein, and Huang}]{XuSGGH23}
Yuancheng Xu, Yanchao Sun, Micah Goldblum, Tom Goldstein, and Furong Huang.
  2023.
\newblock \href {https://doi.org/10.48550/arXiv.2302.03015} {Exploring and
  exploiting decision boundary dynamics for adversarial robustness}.
\newblock \emph{CoRR}, abs/2302.03015.

\bibitem[{Yang et~al.(2020)Yang, Chen, Hsieh, Wang, and Jordan}]{YangCHWJ20}
Puyudi Yang, Jianbo Chen, Cho{-}Jui Hsieh, Jane{-}Ling Wang, and Michael~I.
  Jordan. 2020.
\newblock \href {http://jmlr.org/papers/v21/19-569.html} {Greedy attack and
  gumbel attack: Generating adversarial examples for discrete data}.
\newblock \emph{J. Mach. Learn. Res.}, 21:43:1--43:36.

\bibitem[{Yang et~al.(2022)Yang, Wang, and He}]{YangWH22}
Yichen Yang, Xiaosen Wang, and Kun He. 2022.
\newblock Robust textual embedding against word-level adversarial attacks.
\newblock In \emph{{UAI}}, volume 180 of \emph{Proceedings of Machine Learning
  Research}, pages 2214--2224. {PMLR}.

\bibitem[{Zang et~al.(2020)Zang, Qi, Yang, Liu, Zhang, Liu, and
  Sun}]{ZangQYLZLS20}
Yuan Zang, Fanchao Qi, Chenghao Yang, Zhiyuan Liu, Meng Zhang, Qun Liu, and
  Maosong Sun. 2020.
\newblock \href {https://doi.org/10.18653/v1/2020.acl-main.540} {Word-level
  textual adversarial attacking as combinatorial optimization}.
\newblock In \emph{ACL'20: Proc. of the 58th Annual Meeting of the Association
  for Computational Linguistics Conference}, pages 6066--6080. Association for
  Computational Linguistics.

\bibitem[{Zeng et~al.(2021)Zeng, Qi, Zhou, Zhang, Ma, Hou, Zang, Liu, and
  Sun}]{ZengQZZMHZLS21}
Guoyang Zeng, Fanchao Qi, Qianrui Zhou, Tingji Zhang, Zixian Ma, Bairu Hou,
  Yuan Zang, Zhiyuan Liu, and Maosong Sun. 2021.
\newblock \href {https://doi.org/10.18653/V1/2021.ACL-DEMO.43} {Openattack: An
  open-source textual adversarial attack toolkit}.
\newblock In \emph{Proceedings of the Joint Conference of the 59th Annual
  Meeting of the Association for Computational Linguistics and the 11th
  International Joint Conference on Natural Language Processing, {ACL} 2021 -
  System Demonstrations, Online, August 1-6, 2021}, pages 363--371. Association
  for Computational Linguistics.

\bibitem[{Zhang et~al.(2015)Zhang, Zhao, and LeCun}]{ZhangZL15}
Xiang Zhang, Junbo~Jake Zhao, and Yann LeCun. 2015.
\newblock \href
  {https://proceedings.neurips.cc/paper/2015/hash/250cf8b51c773f3f8dc8b4be867a9a02-Abstract.html}
  {Character-level convolutional networks for text classification}.
\newblock In \emph{Advances in Neural Information Processing Systems 28: Annual
  Conference on Neural Information Processing Systems 2015, December 7-12,
  2015, Montreal, Quebec, Canada}, pages 649--657.

\bibitem[{Zhang et~al.(2021)Zhang, Zhang, Chen, and He}]{ZhangZC020}
Xinze Zhang, Junzhe Zhang, Zhenhua Chen, and Kun He. 2021.
\newblock \href {https://doi.org/10.18653/v1/2021.acl-long.153} {Crafting
  adversarial examples for neural machine translation}.
\newblock In \emph{ACL-IJCNLP'21: Proc. of the 59th Annual Meeting of the
  Association for Computational Linguistics and the 11th International Joint
  Conference on Natural Language Processing}, pages 1967--1977. Association for
  Computational Linguistics.

\bibitem[{Zhao et~al.(2018)Zhao, Dua, and Singh}]{ZhaoDS18}
Zhengli Zhao, Dheeru Dua, and Sameer Singh. 2018.
\newblock \href {https://openreview.net/forum?id=H1BLjgZCb} {Generating natural
  adversarial examples}.
\newblock In \emph{ICLR'18: Proc. of the 6th International Conference on
  Learning Representations}. OpenReview.net.

\bibitem[{Zhou et~al.(2019)Zhou, Jiang, Chang, and Wang}]{ZhouJCW19}
Yichao Zhou, Jyun{-}Yu Jiang, Kai{-}Wei Chang, and Wei Wang. 2019.
\newblock \href {https://doi.org/10.18653/v1/D19-1496} {Learning to
  discriminate perturbations for blocking adversarial attacks in text
  classification}.
\newblock In \emph{EMNLP-IJCNLP'19: Proc. of the Conference on Empirical
  Methods in Natural Language Processing and the 9th International Joint
  Conference on Natural Language Processing}, pages 4903--4912. Association for
  Computational Linguistics.

\bibitem[{Zhu et~al.(2020)Zhu, Cheng, Gan, Sun, Goldstein, and
  Liu}]{ZhuCGSGL20}
Chen Zhu, Yu~Cheng, Zhe Gan, Siqi Sun, Tom Goldstein, and Jingjing Liu. 2020.
\newblock \href {https://openreview.net/forum?id=BygzbyHFvB} {Freelb: Enhanced
  adversarial training for natural language understanding}.
\newblock In \emph{ICLR'20: Proc. of the 8th International Conference on
  Learning Representations}. OpenReview.net.

\end{thebibliography}
\bibliographystyle{acl_natbib}

\section{Reproducibility}
To encourage everyone interested in our work to implement \our, we have taken the following steps:
\begin{itemize}
\item We have created an online click-to-run demo alailable at \url{https://tinyurl.com/22ercuf8} for easy evaluation. Everyone can input adversarial examples and obtain the repaired examples immediately.

\item We have released the detailed source codes and processed datasets that can be retrieved in the supplementary materials. This enables everyone to access the official implementation, aiding in understanding the paper and facilitating their own implementations.

\item We will also release an online benchmark tool for evaluating the performance of adversarial attackers under the defense of \our. This step is essential for reducing evaluation variance across different codebases.
\end{itemize}
These efforts are aimed at promoting the reproducibility of our work and facilitating its implementation by the research community.

\appendix
\section{Adversarial Attack}
\label{app:preliminary}

\subsection{Word-level Adversarial Attack}

Our focus is on defending against word-level adversarial attacks. However, our method can be easily adapted to different types of adversarial attacks. Let $x=\left(x_1, x_2, \cdots, x_n\right)$ be a natural sentence, where $x_i$, $1 \leq i \leq n$, denotes a word. $y$ is the ground truth label. Word-level attackers generally replace some original words with similar words (e.g., synonyms) to fool the objective model. For example, substituting $x_i$ with $\hat{x}_i$ generates an adversary: $\hat{x}=\left(x_1, \cdots, \hat{x}_i, \cdots, x_n\right)$, where $\hat{x}_i$ is an alternative substitution for $x_i$. For an adversary $\hat{x}$, the objective model $F$ predicts its label as follows:
\begin{equation}
\hat{y}=\operatorname{argmax}F\left(\cdot |\hat{x}\right),
\end{equation}
where $\hat{y} \neq y$ if $\hat{x}$ is a successful adversary. To represent adversarial attacks to $F$ using an adversarial attacker $\mathcal{A}$, we denote an adversarial attack as follows:
\begin{equation}
(\hat{x}, \hat{y})\leftarrow \mathcal{A}(F, (x, y)),
\label{eq:attack}
\end{equation}
where $x$ and $y$ denote the natural example and its true label. $\hat{x}$ and $\hat{y}$ are the perturbed adversary and label, respectively.

\subsection{Investigation of Textual Adversarial Attack}
This section delves into an examination of textual adversarial attacks. 

Traditional approaches, such as those noted by~\citet{LiJDLW19} and~\citet{EbrahimiRLD18}, often involve character-level modifications to words (e.g., changing "good" to "go0d") to deceive models by altering their statistical patterns.
In a different approach, knowledge-based perturbations, exemplified by the work of~\citet{ZangQYLZLS20}, employ resources like HowNet to confine the search space, especially in terms of substituting words.

Recent research~\citep{GargR20,LiMGXQ20} has investigated using pre-trained models for generating context-aware perturbations~\citep{LiZPCBSD21}. Semantic-based methods, such as SemAttack~\citep{WangXLCL22}, typically use BERT embedding clusters to create sophisticated adversarial examples.
This differs from prior heuristic methods that employed greedy algorithms~\citep{YangCHWJ20,JinJZS20} or genetic algorithms~\citep{AlzantotSEHSC18,ZangQYLZLS20}, as well as gradient-based techniques~\citep{WangPPCWL20,GuoSJK21} that concentrated on syntactic limitations.

With the evolution of adversarial attack techniques, numerous tools such as TextAttack~\citep{MorrisLYGJQ20} and OpenAttack~\citep{ZengQZZMHZLS21} have been developed and made available in the open-source community. 
These resources facilitate deep learning researchers to efficiently assess adversarial robustness with minimal coding. 
Therefore, our experiments in adversarial defense are conducted using the TextAttack framework, and we extend our gratitude to the authors and contributors of TextAttack for their significant efforts.

\section{Experiments Implementation}
\label{app:experiments}


\subsection{Experimental Adversarial Attackers}
\label{app:sampling-attackers}
We employ \bae, \pwws, and \textfooler\ to generate adversaries for training the adversarial detector. These attackers are chosen because they represent different types of attacks, allowing us to train a detector capable of recognizing a variety of adversarial attacks. This detector exhibits good generalization ability, which we confirm through experiments with other adversarial attackers such as \iga, \wordbugger, \pso, and \clare. Including a larger number of adversarial attackers in the training process can further enhance the performance of the detector. We provide a brief introduction to these adversarial attackers:
\begin{enumerate}[leftmargin=*,label=\alph*),nosep,noitemsep,nolistsep]
\item \textbf{\bae}~\citep{GargR20} generates perturbations by replacing and inserting tagged words based on the candidate words generated by the masked language model (MLM). To identify the most important words in the text, \bae\ employs a word deletion-based importance evaluation method.
\item \textbf{\pwws}~\citep{RenDHC19} is an adversarial attacker based on synonym replacement, which combines word significance and classification probability for word replacement.
\item \textbf{\textfooler}~\citep{JinJZS20} considers additional constraints (such as prediction consistency, semantic similarity, and fluency) when generating adversaries. \textfooler\ uses a gradient-based word importance measure to locate and perturb important words.
\end{enumerate}

\subsection{Hyperparameter Settings}
\label{app:hyperparameter}

We employ the following configurations for fine-tuning classifiers:
\begin{enumerate}[leftmargin=*,noitemsep,nolistsep]
\item The learning rates for both \bert~ and \deberta~ are set to $2 \times 10^{-5}$.
\item The batch size is $16$, and the maximum sequence modeling length is $128$.
\item Dropouts are set to $0.1$ for all models.
\item The loss functions of all objectives use cross-entropy.
\item The victim models and \our~models are trained for $5$ epochs.
\item The optimizer used for fine-tuning objective models is \texttt{AdamW}.
\end{enumerate}
Please refer to our released code for more details.

\subsection{Evaluation Metrics}
\label{app:metrics}
In this section, we introduce the adversarial defense metrics. First, we select a target dataset, referred to as $\mathcal{D}$, containing only natural examples. Our goal is to generate adversaries that can deceive a victim model $F_{J}$. We group the successful adversaries into a subset called $\mathcal{D}_{adv}$ and the remaining natural examples with no adversaries into another subset called $\mathcal{D}_{nat}$. We then combine these two subsets to form the attacked dataset, $\mathcal{D}_{att}$. We apply \our\ to $\mathcal{D}_{att}$ to obtain the repaired dataset, $\mathcal{D}_{rep}$. The evaluation metrics used in the experiments are described as follows:
\begin{equation*}
    \textsc{NtA}~=\frac{TP_{\mathcal{D}}+TN_{\mathcal{D}}}{P_{\mathcal{D}}+N_{\mathcal{D}}}
\end{equation*}
\begin{equation*}
    \textsc{AtA}~=\frac{TP_{\mathcal{D}_{att}}+TN_{\mathcal{D}_{att}}}{P_{\mathcal{D}_{att}}+N_{\mathcal{D}_{att}}}
\end{equation*}
\begin{equation*}
    \textsc{DtA}~=\frac{TP_{\mathcal{D}_{adv}}^{*}+TN_{\mathcal{D}_{adv}}^{*}}{P_{\mathcal{D}_{adv}}^{*}+N_{\mathcal{D}_{adv}}^{*}}
\end{equation*}
\begin{equation*}
    \textsc{DfA}~=\frac{TP_{\mathcal{D}_{adv}}+TN_{\mathcal{D}_{adv}}}{P_{\mathcal{D}_{adv}}+N_{\mathcal{D}_{adv}}}
\end{equation*}
\begin{equation*}
    \textsc{RpA}~=\frac{TP_{\mathcal{D}_{rep}}+TN_{\mathcal{D}_{rep}}}{P_{\mathcal{D}_{rep}}+N_{\mathcal{D}_{rep}}}
\end{equation*}
where $TP$ ,$TN$, $P$ and $N$ are the number of true positives and true negatives, positive and negative in standard classification, respectively. $TP^{*}$, $TN^{*}$, $P^{*}$ and $N^{*}$ indicate the case numbers in adversarial detection.

\subsection{Experimental Environment}
\label{app:environment}
The experiments are carried out on a computer running the Cent OS 7 operating system, equipped with an RTX 3090 GPU and a Core i-12900k processor. We use the PyTorch 1.12 library and a modified version of TextAttack, based on version 0.3.7.

\section{Ablation Experiments}

\subsection{Defense of LLM-based Adversarial Attack}

\begin{table}[htbp]
	\centering
	\caption{Defense performance of \our\ against adversarial attacks generated by \texttt{ChatGPT-3.5}.}
	\resizebox{.8\linewidth}{!}{
		\begin{tabular}{c|c|c|c}
			\hline
			\multirow{1}[1]{*}{\textsc{Dataset}} & \multirow{1}[1]{*}{\textsc{Attacker}}  & \textsc{DfA}  & \textsc{RpA}  \\
   			\hline
			\multirow{2}{*}{\agnews} & \multirow{2}{*}{\textsc{ChatGPT}} & \rsv & $59.0$ \\
			& & \our & $\mathbf{72.0}$ \\
			\hline
   			\multirow{2}{*}{\yahoo} & \multirow{2}{*}{\textsc{ChatGPT}} & \rsv & $49.0$ \\
			& & \our & $\mathbf{61.0}$ \\
			\hline
			\multirow{2}{*}{\sst2} & \multirow{2}{*}{\textsc{ChatGPT}} & \rsv & $37.0$ \\
			& & \our & $\mathbf{74.0}$ \\
			\hline
			\multirow{2}{*}{\amazon} & \multirow{2}{*}{\textsc{ChatGPT}} & \rsv & $58.0$ \\
			& & \our & $\mathbf{}{82.0}$ \\
			\hline
		\end{tabular}
	}
	\label{tab:llm}
\end{table}

Recent years have witnessed the superpower of large language models (LLMs) such as \texttt{ChatGPT}~\citep{OpenAI23}, 
which we hypothesize to have a stronger ability to generate adversaries. In this subsection, we evaluate the defense performance of \our\ against adversaries generated by \texttt{ChatGPT-3.5}. 
Specifically, for each dataset considered in our previous experiments, we use \texttt{ChatGPT}\footnote{\texttt{ChatGPT3.5-0301}} to generate $100$ adversaries and investigate the defense accuracy achieved by \our. 

From the experimental results shown in~\pref{tab:llm}, we find that \our\ consistently outperforms \rsv\ in terms of defense accuracy. 
Specifically, in the \sst2\ dataset, \rsv\ records a defense accuracy of 37.0\%, however, \our\ impressively repairs 74.0\% of the attacks. 
Similar trends hold for the \amazon\ and \agnews\ datasets, where \our\ achieves defense accuracy of 82.0\% and 72.0\% respectively, in contrast to the 58.0\% and 59.0\% offered by \rsv. 
In conclusion, \our\ can defend against various unknown adversarial attacks which have a remarkable performance in contrast to existing adversarial defense approaches.

\subsection{Performance of \our\ based on Different $\hat{\mathcal{A}}_{PD}$}

In \our, \pd\ can incorporate any adversarial attacker or even an ensemble of attackers, as the process doesn't require prior knowledge of the specific malicious perturbations. 
Regardless of which adversaries are deployed against \our, \pwws\ consistently seeks safe perturbations for the current adversarial examples. 
The abstract nature of \pd\ is critical, allowing for adaptability and effectiveness against a broad spectrum of adversarial attacks, rendering it a versatile defense mechanism in our study.

In order to investigate the impact of $\hat{\mathcal{A}}_{PD}$ in \pb, we have implemented further experiments to demonstrate the adversarial defense performance of \pd\ using different attackers, e.g., \textfooler\ and \bae. 
The results are shown in \pref{tab:backend_ablations}. According to the experimental results, it is observed that \pwws\ has a similar performance to \textfooler\ in \pd, 
while \bae\ is slightly inferior to both \pwws\ and \textfooler. 
However, the variance are not significant among different attackers in \pd, which means the performance of \our\ is not sensitive to the choice of $\hat{\mathcal{A}}_{PD}$, in contrast to the adversarial attack performance of the adversarial attacker.
\begin{table*}[hbtp]
		\centering
		\caption{The adversarial detection and defense performance of \our\ based on different backends ($\hat{\mathcal{A}}_{PD}$).  We report the average accuracy of five random runs. ``\textsc{TF}'' indicates \textfooler.}
		\resizebox{\linewidth}{!}{
        \renewcommand{\arraystretch}{1.3}
            \begin{tabular}{c@{\hspace{0.1cm}}|c@{\hspace{0.1cm}}|c@{\hspace{0.02cm}}c@{\hspace{0.02cm}}c@{\hspace{0.02cm}}c@{\hspace{0.02cm}}c@{\hspace{0.02cm}}|c@{\hspace{0.02cm}}c@{\hspace{0.02cm}}c@{\hspace{0.02cm}}c@{\hspace{0.02cm}}c@{\hspace{0.02cm}}|c@{\hspace{0.02cm}}c@{\hspace{0.02cm}}c@{\hspace{0.02cm}}c@{\hspace{0.02cm}}c@{\hspace{0.02cm}}|c@{\hspace{0.02cm}}c@{\hspace{0.02cm}}c@{\hspace{0.02cm}}c@{\hspace{0.02cm}}c@{\hspace{0.02cm}}}
    \hline
    \multirow{2}[1]{*}{\textsc{Defender}} & \multirow{2}[1]{*}{\textsc{Attacker}} & \multicolumn{5}{c}{\agnews (4-category)}            & \multicolumn{5}{|c}{\yahoo (10-category)}            & \multicolumn{5}{|c}{\sst2 (2-category)}            & \multicolumn{5}{|c}{\amazon (2-category)} \\
\cline{3-22}          &       & \multicolumn{1}{c}{\textsc{NtA}} & \multicolumn{1}{c}{\textsc{AtA}} & \multicolumn{1}{c}{\textsc{DtA}} & \multicolumn{1}{c}{\textsc{DfA}} & \multicolumn{1}{c}{\textsc{RpA}} & \multicolumn{1}{|c}{\textsc{NtA}} & \multicolumn{1}{c}{\textsc{AtA}} & \multicolumn{1}{c}{\textsc{DtA}} & \multicolumn{1}{c}{\textsc{DtA}} & \multicolumn{1}{c}{\textsc{RpA}} & \multicolumn{1}{|c}{\textsc{NtA}} & \multicolumn{1}{c}{\textsc{AtA}} & \multicolumn{1}{c}{\textsc{DtA}} & \multicolumn{1}{c}{\textsc{DfA}} & \multicolumn{1}{c}{\textsc{RpA}} & \multicolumn{1}{|c}{\textsc{NtA}} & \multicolumn{1}{c}{\textsc{AtA}} & \multicolumn{1}{c}{\textsc{DtA}} & \multicolumn{1}{c}{\textsc{DfA}} & \multicolumn{1}{c}{\textsc{RpA}} \\
    \hline
    \hline
\multirow{3}[2]{*}{\our~(\pwws)} 
          & \pwws          &       & $32.09$ & $90.11$ & $95.88$ & $92.36$ &       & $5.70$  & $87.33$ & $92.47$ & $69.40$ &       & $23.44$ & $94.03$ & $98.62$ & $89.85$ &       & $15.56$ & $97.33$ & $99.99$ & $94.42$ \\
          & \textsc{TF}    & $94.30$ & $50.50$ & $90.29$ & $96.76$ & $92.14$ & $76.45$ & $13.60$ & $87.49$ & $93.54$ & $70.50$ & $91.55$ & $16.21$ & $94.03$ & $99.86$ & $89.72$ & $94.32$ & $21.77$ & $93.85$ & $99.99$ & $93.96$ \\
          & \bae           &       & $74.80$ & $57.55$ & $96.25$ & $93.64$ &       & $27.50$ & $82.46$ & $96.30$ & $73.06$ &       & $35.21$ & $78.99$ & $99.28$ & $89.77$ &       & $44.00$ & $80.55$ & $99.99$ & $93.89$ \\
    \hline
    \hline
    \multirow{3}[2]{*}{\our~(\texttt{TF})} 
          & \pwws          &       & $32.09$ & $83.67$ & $94.07$ & $92.27$ &       & $5.70$  & $65.01$ & $83.25$ & $65.33$ &       & $23.44$ & $36.90$ & $98.90$ & $90.67$ &       & $15.56$ & $29.60$ & $99.99$ & $94.33$ \\
          & \textsc{TF}    & $94.30$ & $50.50$ & $82.44$ & $96.46$ & $92.67$ & $76.45$ & $13.60$ & $74.21$ & $92.96$ & $71.00$ & $91.55$ & $16.21$ & $39.70$ & $99.98$ & $90.73$ & $94.32$ & $21.77$ & $40.70$ & $99.99$ & $94.33$ \\
          & \bae           &       & $74.80$ & $46.98$ & $92.68$ & $91.00$ &       & $27.50$ & $37.41$ & $86.49$ & $72.67$ &       & $35.21$ & $19.84$ & $99.98$ & $91.33$ &       & $44.00$ & $38.59$ & $99.99$ & $94.33$ \\
    \hline
    \hline
    \multirow{3}[2]{*}{\our~(\bae)} 
          & \pwws          &       & $32.09$ & $83.67$ & $93.22$ & $92.08$ &       & $5.70$  & $65.01$ & $81.15$ & $64.00$ &       & $23.44$ & $36.90$ & $93.92$ & $87.67$ &       & $15.56$ & $29.60$ & $99.54$ & $94.00$ \\
          & \textsc{TF}    & $94.30$ & $50.50$ & $82.44$ & $95.96$ & $92.33$ & $76.45$ & $13.60$ & $74.21$ & $87.79$ & $67.33$ & $91.55$ & $16.21$ & $39.70$ & $96.55$ & $89.00$ & $94.32$ & $21.77$ & $40.70$ & $99.61$ & $93.64$ \\
          & \bae           &       & $74.80$ & $46.98$ & $95.12$ & $91.33$ &       & $27.50$ & $37.41$ & $83.78$ & $72.00$ &       & $35.21$ & $19.84$ & $97.55$ & $90.00$ &       & $44.00$ & $38.59$ & $99.15$ & $93.80$ \\

    \hline
    \end{tabular}%
		}
		\label{tab:backend_ablations}
\end{table*}

\subsection{Performance of \our\ without Adversarial Training Objective}

The rationale behind the adversarial training objective $\mathcal{L}_a$ in our study is founded on two key hypotheses. 
\begin{enumerate}[leftmargin=*,label=\alph*),nosep,noitemsep,nolistsep]
    \item \textbf{Enhancing Adversarial Detection:} We recognize an implicit link between the tasks of adversarial training and adversarial example detection. 
    Our theory suggests that by incorporating an adversarial training objective, we can indirectly heighten the model's sensitivity to adversarial examples, leading to more accurate detection of such instances.
    \item \textbf{Improving Model Robustness:} We posit that an adversarial training objective can bolster the model's robustness, thereby mitigating performance degradation when the model faces an attack. 
    This approach is designed to strengthen the model against potential adversarial threats.    
\end{enumerate}

To validate these hypotheses, we conducted ablation experiments on the adversarial training objective. 
The experimental setup was aligned with that described in \pref{tab:main}, and the results are outlined in \pref{tab:backend_at}.

\begin{table*}[hbtp]
    \centering
    \caption{The adversarial detection and defense performance of \our\ with (``w/'') and without (``w/o'') the adversarial training objective.  We report the average accuracy of five random runs. ``\textsc{TF}'' indicates \textfooler.}
    \resizebox{\linewidth}{!}{
    \renewcommand{\arraystretch}{1.3}
        \begin{tabular}{c@{\hspace{0.1cm}}|c@{\hspace{0.1cm}}|c@{\hspace{0.02cm}}c@{\hspace{0.02cm}}c@{\hspace{0.02cm}}c@{\hspace{0.02cm}}c@{\hspace{0.02cm}}|c@{\hspace{0.02cm}}c@{\hspace{0.02cm}}c@{\hspace{0.02cm}}c@{\hspace{0.02cm}}c@{\hspace{0.02cm}}|c@{\hspace{0.02cm}}c@{\hspace{0.02cm}}c@{\hspace{0.02cm}}c@{\hspace{0.02cm}}c@{\hspace{0.02cm}}|c@{\hspace{0.02cm}}c@{\hspace{0.02cm}}c@{\hspace{0.02cm}}c@{\hspace{0.02cm}}c@{\hspace{0.02cm}}}
\hline
    \multirow{2}[1]{*}{\textsc{Defender}} & \multirow{2}[1]{*}{\textsc{Attacker}} & \multicolumn{5}{c}{\agnews (4-category)}            & \multicolumn{5}{|c}{\yahoo (10-category)}            & \multicolumn{5}{|c}{\sst2 (2-category)}            & \multicolumn{5}{|c}{\amazon (2-category)} \\
\cline{3-22}          &       & \multicolumn{1}{c}{\textsc{NtA}} & \multicolumn{1}{c}{\textsc{AtA}} & \multicolumn{1}{c}{\textsc{DtA}} & \multicolumn{1}{c}{\textsc{DfA}} & \multicolumn{1}{c}{\textsc{RpA}} & \multicolumn{1}{|c}{\textsc{NtA}} & \multicolumn{1}{c}{\textsc{AtA}} & \multicolumn{1}{c}{\textsc{DtA}} & \multicolumn{1}{c}{\textsc{DtA}} & \multicolumn{1}{c}{\textsc{RpA}} & \multicolumn{1}{|c}{\textsc{NtA}} & \multicolumn{1}{c}{\textsc{AtA}} & \multicolumn{1}{c}{\textsc{DtA}} & \multicolumn{1}{c}{\textsc{DfA}} & \multicolumn{1}{c}{\textsc{RpA}} & \multicolumn{1}{|c}{\textsc{NtA}} & \multicolumn{1}{c}{\textsc{AtA}} & \multicolumn{1}{c}{\textsc{DtA}} & \multicolumn{1}{c}{\textsc{DfA}} & \multicolumn{1}{c}{\textsc{RpA}} \\
      \hline
      \hline
\multirow{3}[2]{*}{\our~(w/ $\mathcal{L}_a$)} 
            & \pwws          &       & $32.09$ & $90.11$ & $95.88$ & $92.36$ &       & $5.70$  & $87.33$ & $92.47$ & $69.40$ &       & $23.44$ & $94.03$ & $98.62$ & $89.85$ &       & $15.56$ & $97.33$ & $99.99$ & $94.42$ \\
            & \textsc{TF}    & $94.30$ & $50.50$ & $90.29$ & $96.76$ & $92.14$ & $76.45$ & $13.60$ & $87.49$ & $93.54$ & $70.50$ & $91.55$ & $16.21$ & $94.03$ & $99.86$ & $89.72$ & $94.32$ & $21.77$ & $93.85$ & $99.99$ & $93.96$ \\
            & \bae           &       & $74.80$ & $57.55$ & $96.25$ & $93.64$ &       & $27.50$ & $82.46$ & $96.30$ & $73.06$ &       & $35.21$ & $78.99$ & $99.28$ & $89.77$ &       & $44.00$ & $80.55$ & $99.99$ & $93.89$ \\
      \hline
      \hline
      \multirow{3}[2]{*}{\our~(w/o $\mathcal{L}_a$)} 
            & \pwws          &       & $11.10$ & $82.88$ & $92.07$ & $90.70$ &       & $3.46$ & $78.43$ & $87.42$ & $63.79$ &       & $10.70$ & $91.41$ & $99.62$ & $89.60$ &       & $16.5$ & $96.50$ & $99.30$ & $93.60$ \\
            & \textsc{TF}    & $94.44$ & $16.09$ & $84.88$ & $93.07$ & $87.28$ & $76.32$ & $0.42$ & $78.65$ & $78.36$ & $56.72$ & $91.54$ & $5.30$ & $89.48$ & $95.15$ & $85.80$ & $94.29$ & $17.53$ & $98.63$ & $99.17$ & $92.78$ \\
            & \bae           &       & $67.93$ & $83.17$ & $91.49$ & $91.15$ &       & $45.10$ & $71.89$ & $75.47$ & $64.56$ &       & $25.70$ & $57.01$ & $95.64$ & $87.10$ &       & $45.54$ & $92.67$ & $99.48$ & $93.31$ \\

      \hline
      \end{tabular}%
          }
      \label{tab:backend_at}
  \end{table*}
  
  These experimental findings reveal that omitting the adversarial training objective in \our\ consistently leads to a reduction in model robustness across all datasets. 
  This reduction can be as substantial as approximately 30\%, adversely affecting the performance of the adversarial defense. 
  Additionally, adversarial detection capabilities also diminish, with the most significant drop being around 20\%. 
  These results highlight the critical role of the adversarial training objective in \our, confirming its efficacy in enhancing both model robustness and adversarial example detection capabilities.

\subsection{Performance of \our\ without Multitask Training Objective}

\begin{table}[ht]
    \centering
    \resizebox{.8\linewidth}{!}{
    \begin{tabular}{cccc}
    \hline
    \textsc{Dataset} & \textsc{Model} & \textsc{Victim-S} & \textsc{Victim-M}  \\ 
    \hline
    \agnews & \bert & $94.30$ & $93.90$ ($-0.40\downarrow$) \\
    \yahoo  & \bert & $76.45$ & $76.61$ ($+0.16\uparrow$) \\
    \sst2   & \bert & $91.70$ & $91.49$ ($-0.21\downarrow$) \\
    \amazon & \bert & $94.24$ & $94.24$ (---) \\
    \hline
    \end{tabular}
    }
    \caption{Victim model's accuracy (\%) on clean dataset-based single-task and multitask training scenarios, i.e., \textbf{Victim-S} and \textbf{Victim-M} respectively. The experiments are based on the \bert\ model. }
    \label{tab:ablation_mt}
\end{table}

Before developing \our, we carefully considered the potential impact on classification performance due to multitask training objectives. This consideration was explored in our proof-of-concept experiments.

To delve deeper into this impact, we trained victim models as single-task models (i.e., no adversarial detection objective and adversarial training objective), instead of multitask training, and then collated detailed results for comparison with \our. 
In this experiment, we focused solely on evaluating performance using pure natural examples. 
The results of this comparison are outlined in \pref{tab:ablation_mt}. The symbols "$\uparrow$" and "$\downarrow$" accompanying the numbers indicate whether the performance is better or worse than that of the single-task model, respectively.

Based on these results, it is apparent that the inclusion of additional loss terms in multitask training objectives does impact the victim model's performance on clean examples. 
However, this influence is not substantial across all datasets and shows only slight variations. This finding suggests that the impact of multitask training objectives is relatively minor when compared to traditional adversarial training methods.

\subsection{Performance Comparison between \our\ and Adversarial Training Baseline}

\begin{table}[htbp]
    \centering
    \resizebox{0.7\linewidth}{!}{
    \begin{tabular}{cccc}
    \hline
    \textsc{Dataset} & \textsc{Attacker} & \our & \textsc{AT}  \\
    \hline
    & \pwws    & $92.36$  & $60.10$                 \\ 
    \agnews  & \textsc{TF}     & $92.14$ & $61.87$          \\ 
            & \bae     & $93.64$  & $63.62$                 \\ 
            \hline
            & \pwws    & $69.40$  & $40.21$                \\ 
    \yahoo  & \textsc{TF}     & $70.50$ & $38.75$          \\ 
            & \bae     & $73.06$  & $42.97$                \\ 
            \hline
            & \pwws    & $89.85$  & $32.46$                \\ 
    \sst2    & \textsc{TF}     & $89.72$ & $31.23$          \\ 
            & \bae     & $89.77$  & $34.61$              \\ 
            \hline
            & \pwws    & $94.42$  & $51.90$                \\ 
    \amazon  & \textsc{TF}     & $93.96$ & $49.49$          \\ 
            & \bae     & $93.89$  & $49.75$                \\ 

    \hline    
    \end{tabular}
    }
    \caption{The repaired performance of \our\ and the adversarial training baseline.  We report the average accuracy of five random runs. ``\textsc{TF}'' indicates \textfooler.}
    \label{tab:baseline_at}
    \end{table}
    
We have conducted experiments to showcase the experimental results of the adversarial training baseline (\textsc{AT}). 
The victim model is \bert, and the experimental setup is the same as for \our, including the number of adversaries used for training. 
We only show the metric of repaired accuracy, as \textsc{AT} does not support detect-to-defense. The results (i.e., \textsc{RpA} (\%)) are available in \pref{tab:baseline_at}.

For these experiments, we used \bert\ as the victim model and maintained the same experimental setup as for \our, including the number of adversaries used for training. 
It's important to note that we focus solely on the repaired accuracy metric, as \textsc{AT} does not facilitate detect-to-defense functionality. 
From these results, it becomes apparent that the traditional adversarial training baseline is less effective compared to \our, which utilizes perturbation defocusing. 
Specifically, the adversarial defense accuracy of \textsc{AT} is generally below 50\%. 
This finding underscores the limitations of traditional adversarial training methods, particularly their high cost and reduced effectiveness against adapted adversarial attacks.

\subsection{Efficiency Evaluation of \our}
The main efficiency depends on multiple adversarial perturbations search. We have implemented two experiments to investigate the efficiency of \our. 
Please note that the time costs for adversarial attack and defense are dependent on specific software and hardware environments. 

\textbf{Time Costs for Multiple Examples}. We have collected three small sub-datasets that contain different numbers of adversarial examples and natural examples, say 200:0, 100:100, and 0:200. 
We apply adversarial detection and defense to this dataset and calculate the time costs. The results (measurement: second) are available in \pref{tab:efficiency1}.

\begin{table*}[t!]
    \centering
    \resizebox*{\linewidth}{!}{
    \begin{tabular}{c|ccc|ccc|ccc|ccc}
    \hline
    \multirow{2}{*}{\textsc{Attacker}} & \multicolumn{3}{c|}{\agnews} & \multicolumn{3}{c|}{\yahoo} & \multicolumn{3}{c|}{\sst2} & \multicolumn{3}{c}{\amazon} \\ \cline{2-13} 
                            & 200:0 & 100:100 & \multicolumn{1}{c|}{0:200} & 200:0 & 100:100 & \multicolumn{1}{c|}{0:200} & 200:0 & 100:100 & \multicolumn{1}{c|}{0:200} & 200:0 & 100:100 & 0:200 \\ 
    \hline
    \pwws                    &       & $142.090$ & $298.603$ &       & $313.317$ & $621.196$ &       & $36.268$ & $126.054$ &       & $438.532$ & $875.083$ \\ 
    \texttt{TF}              & $1.188$ & $146.654$ & $293.542$ & $1.157$ & $314.926$ & $642.206$ & $1.092$ & $51.303$ & $137.795$ & $1.138$ & $329.075$ & $665.052$ \\ 
    \bae                     &       & $141.434$ & $260.231$ &       & $352.186$ & $876.606$ &       & $52.626$ & $138.325$ &       & $349.256$ & $655.264$ \\ 

    \hline
    \end{tabular}
    }
    \caption{The efficiency of \our\ defending against different adversarial attacks with different portions of natural and adversarial instances. The measurement is second.}
    \label{tab:efficiency1}
\end{table*}

\textbf{Time Costs for Single Examples}. We have also detailed the time costs per natural example, adversarial attack, and adversarial defense in \pd\. 
The experimental results can be found in \pref{tab:efficiency2}.

\begin{table*}[ht]
    \centering
    \resizebox*{\linewidth}{!}{
    \begin{tabular}{c|c|ccc|ccc|ccc|ccc}
    \hline
    \multirow{2}{*}{\textsc{Defender}} & \multirow{2}{*}{\textsc{Attacker}} & \multicolumn{3}{c|}{\agnews} & \multicolumn{3}{c|}{\yahoo} & \multicolumn{3}{c|}{\sst2} & \multicolumn{3}{c}{\amazon} \\
     \cline{3-14}             &           & \textsc{Clean}     & \textsc{Attack}  & \textsc{Defense} & \textsc{Clean}     & \textsc{Attack}  & \textsc{Defense} & \textsc{Clean}     & \textsc{Attack}  & \textsc{Defense} & \textsc{Clean}     & \textsc{Attack}  & \textsc{Defense} \\ 
     \hline
                              & PWWS      &           & $2.081$   & $1.356$   &           & $4.958$   & $3.308$   &           & $0.529$   & $0.588$   &           & $4.745$   & $3.678$   \\ 
    \our                      & TF        & $0.008$     & $2.460$   & $1.317$   & $0.008$     & $4.693$   & $3.128$   & $0.006$     & $0.662$   & $0.571$   & $0.007$     & $4.003$   & $4.607$   \\ 
                              & BAE       &           & $2.464$   & $1.295$   &           & $5.194$   & $4.053$   &           & $0.669$   & $0.594$   &           & $4.350$   & $4.403$   \\

    \hline
    \end{tabular}
    }
    \caption{The execution efficiency of inferring clean examples, generating, and defending against adversarial examples.}
    \label{tab:efficiency2}
\end{table*}

According to the experimental results, \pd\ is slightly faster than the adversarial attack in most cases. 
Intuitively, the perturbed semantics in a malicious adversarial example are generally not robust, as most of the deep semantics remain within the adversarial example. 
Therefore, \our is able to rectify the example with fewer perturbations needed to search.

\section{Deployment Demo}
\label{app:demo}
\begin{figure*}
\centering
\includegraphics[width=0.7\linewidth]{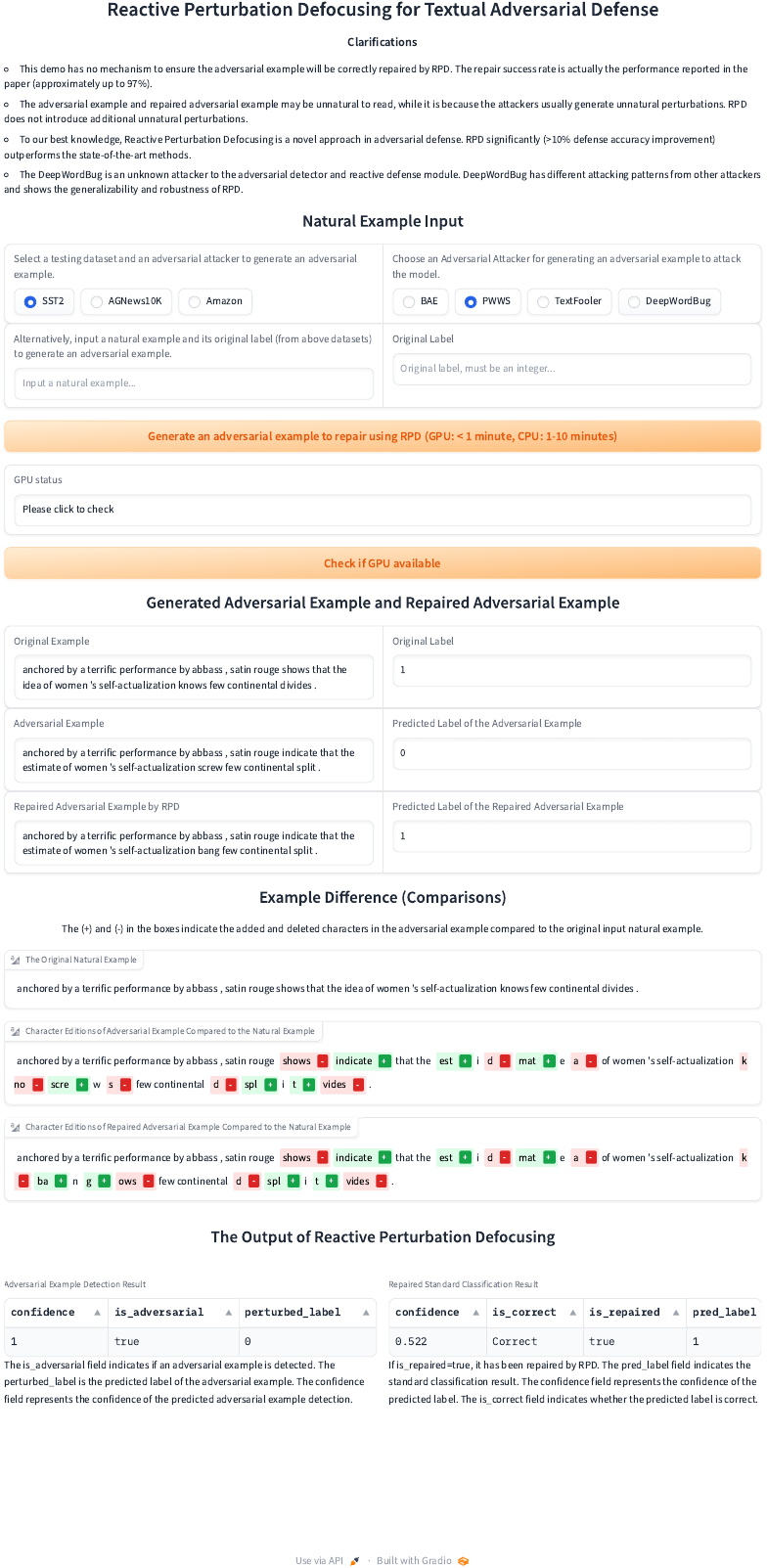}
\caption{The demo examples of adversarial detection and defense built on \our~for defending against multi-attacks. The comparisons between natural and repaired examples are available based on the ``\textit{difflib}'' library. The ``$+$'' and ``$-$'' in the colored boxes indicate letters addition and deletion compared to the natural examples. It is observed that \our\ only injects only one perturbation to repair the adversarial example, i.e., changing ``screw'' to ``bang'' in the adversarial example.}
\label{fig:demo}
\end{figure*}

We have created an anonymous demonstration of \our, which is available on Huggingface Space\footnote{\url{https://huggingface.co/spaces/anonymous8/RPD-Demo}}. To illustrate the usage of our method, we provide two examples in \pref{fig:demo}. In this demonstration, users can either input a new phrase along with a label or randomly select an example from a supplied dataset, to perform an attack, adversarial detection, and adversarial repair.

\end{document}